\documentclass[11pt]{article}

\usepackage[preprint]{acl}

\usepackage{times}
\usepackage{latexsym}

\usepackage[T1]{fontenc}

\usepackage[utf8]{inputenc}

\usepackage{microtype}

\usepackage{inconsolata}

\usepackage{graphicx}

\usepackage{amsmath,amssymb,amsthm,mathtools,bm}
\usepackage{booktabs,array,tabularx,multirow}
\usepackage{algorithm}
\usepackage[noend]{algpseudocode}
\usepackage{xcolor}
\usepackage{colortbl}
\usepackage[nameinlink,capitalise,noabbrev]{cleveref}
\usepackage{url}
\usepackage{wrapfig}
\usepackage{float}
\usepackage{makecell}
\usepackage{pifont}
\usepackage{enumitem}
\usepackage{titletoc}
\usepackage[most]{tcolorbox}

\newcommand{\ours}{CriticOPD}
\newtcolorbox{prompt}[1][]{enhanced,breakable,colback=gray!4,colframe=black!65,coltitle=white,fonttitle=\small\bfseries,fontupper=\small,boxrule=0.5pt,arc=1.5pt,left=5pt,right=5pt,top=4pt,bottom=4pt,before skip=6pt,after skip=8pt,#1}

\theoremstyle{remark}

\theoremstyle{plain}

\Crefname{theorem}{Theorem}{Theorems}
\crefname{proposition}{Proposition}{Propositions}
\Crefname{proposition}{Proposition}{Propositions}
\crefname{lemma}{Lemma}{Lemmas}
\Crefname{lemma}{Lemma}{Lemmas}
\crefname{corollary}{Corollary}{Corollaries}
\Crefname{corollary}{Corollary}{Corollaries}
\crefname{definition}{Definition}{Definitions}
\Crefname{definition}{Definition}{Definitions}
\crefname{assumption}{Assumption}{Assumptions}
\Crefname{assumption}{Assumption}{Assumptions}
\crefname{equation}{Equation}{Equations}
\Crefname{equation}{Equation}{Equations}
\crefname{algorithm}{Algorithm}{Algorithms}
\Crefname{algorithm}{Algorithm}{Algorithms}
\crefname{table}{Table}{Tables}
\Crefname{table}{Table}{Tables}
\crefname{figure}{Figure}{Figures}
\Crefname{figure}{Figure}{Figures}
\crefname{section}{Section}{Sections}
\Crefname{section}{Section}{Sections}

\title{
Recovering Reliable Supervision via Critique-Guided On-Policy Distillation
}

\author{
\textbf{Kaixiang Zhao}$^{1}$ \quad \textbf{Tianrun Yu}$^{1}$ \quad \textbf{Shangzhe Li}$^{2}$ \quad \textbf{Yuxiao Yang}$^{2}$ \\
\textbf{Amanda Hughes}$^{1}$ \quad \textbf{Weitong Zhang}$^{2}$ \quad \textbf{Taylor W. Killian}$^{1}$ \\
$^{1}$Brigham Young University \quad $^{2}$University of North Carolina at Chapel Hill \\
\texttt{kzhao2@byu.edu} \quad \texttt{tkillian@cs.byu.edu}}

\begin{document}
\maketitle


\begin{abstract}
On-policy distillation (OPD) trains a student model on its own rollouts with token-level supervision from a teacher, assuming that the teacher can reach a correct answer from any prefix the student writes. We call this property the recoverability of the teacher and show it fails on incorrect rollouts: the teacher follows student errors, although it recognizes them. We propose CriticOPD, using error recognition to restore recoverability. The teacher critiques an incorrect rollout, allowing the student to rewrite its solution following the error. The resulting output, including the critique as privileged information, is scored by the teacher while the student uses the revision without the critique for training. On two Qwen3 students and six mathematical reasoning benchmarks, CriticOPD improves OPD, with the largest gains on competition problems. Code is available at\newline {\small\urlstyle{same}\href{https://github.com/kzhao5/CriticOPD}{\nolinkurl{github.com/kzhao5/CriticOPD}}}.
\end{abstract}

\section{Introduction}
\label{sec:intro}

On-policy distillation (OPD) trains a student model on its own rollouts with token-level supervision from a teacher \citep{agarwal2024onpolicy, gu2023minillm}. OPD addresses a weakness of offline distillation. In offline distillation, the student trains on sequences the teacher generates, whereas the student at inference continues from prefixes that it generates itself, so the distribution of prefixes in training differs from the distribution of prefixes that the student visits at inference. We call the distance between these two distributions the occupancy mismatch. Under this mismatch, the errors of the student may be underrepresented in training, and once the student errs at inference, it may enter a prefix that it has rarely seen during training, allowing errors to accumulate \citep{ross2011dagger, agarwal2024onpolicy}. OPD reduces the occupancy mismatch by training on sequences that the student generates, with the next-token distribution of the teacher at each prefix as supervision. However, the effectiveness of this supervision in correcting errors depends on the teacher's ability to recover from the student's prefixes. When the teacher follows the student's errors rather than correcting them, its next-token distribution may no longer provide a reliable corrective target.

\begin{figure}[t]
    \centering
    \includegraphics[width=\columnwidth]{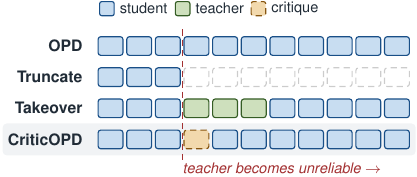}
    \caption{Where the supervision of each method falls after the student errs, and who writes it.}
    \label{fig:schematic}
\end{figure}

\begin{figure*}[t]
    \centering
    \includegraphics[width=\textwidth]{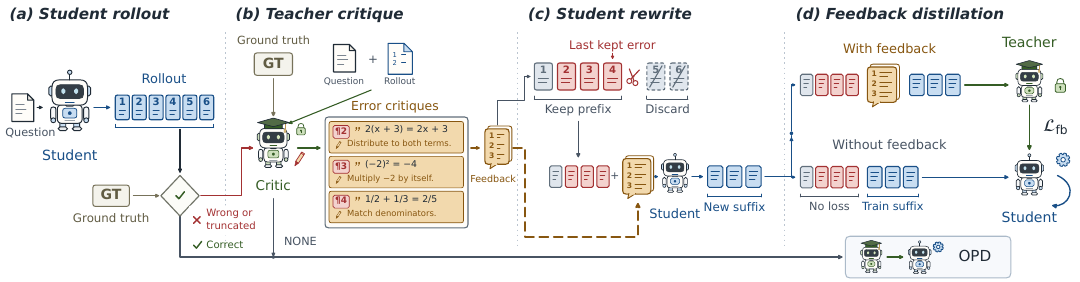}
    \caption{Overview of CriticOPD. The student rewrites an incorrect rollout from the last error that the teacher critiques. The teacher scores the rewrite with the critique, and the student learns from it without the critique.}
    \label{fig:pipeline}
\end{figure*}

Prior work has found that teacher supervision can become less reliable as the prefix of the student grows longer \citep{liu2026sfd, zhou2026early, li2026rethinking}. Existing methods respond in two ways (Figure~\ref{fig:schematic}). The first reduces supervision. It truncates the later part of long responses \citep{yang2026pruneopd, zhang2026fast} or stops rollouts or supervision early \citep{zhou2026early, liu2026prefix}. The second replaces the sequence. It lets the teacher take over a span of
the generation \citep{xu2026relay} or regenerate from an earlier context
that the teacher judges reliable \citep{wang2026backtracking}. Both obtain more reliable supervision by restricting the states receiving supervision or modifying training trajectories, and thereby introduce more occupancy mismatch between training and inference than OPD. Reducing supervision leaves untrained the later states that the student reaches at inference, and replacing the sequence trains the student on sequences that the teacher writes and the student rarely generates. These methods largely avoid unreliable teacher supervision rather than improving its corrective value on erroneous student prefixes. In this work, we aim to make teacher supervision on student-generated continuations more reliable while introducing as little additional occupancy mismatch as possible.

We measure teacher reliability through recoverability, a notion from
imitation learning \citep{swamy2021moments}, which we define as the
teacher's accuracy when continuing from a student prefix. On incorrect rollouts, recoverability declines sharply even on problems the teacher can solve. The teacher often follows the student's errors, with up to half of its failed continuations ending in the student's incorrect answer. Yet its ability to identify errors remains largely intact. This reveals an asymmetry: the teacher may fail to continue correctly but still recognize errors.

Motivated by this observation, we propose CriticOPD. Given an incorrect rollout and the reference answer, the teacher identifies reasoning errors and provides corrective feedback. The student rewrites its solution from the last error using this feedback. The teacher then scores the rewrite with the critique in context, while the student learns without it. The critique serves as privileged information during training \citep{vapnik2009lupi,lopezpaz2016unifying}, enabling the student to learn corrective behavior without feedback at inference. Unlike teacher takeover, CriticOPD keeps all supervised tokens student-generated while improving the recoverability of the revised trajectories.

\section{Measuring the Reliability of Teacher Supervision}
\label{sec:diagnosis}

This section measures the reliability of teacher supervision through recoverability and examines where and why OPD supervision becomes unreliable. Let $h_t=(x,y_{<t})$ denote a student-generated prefix, and let $r(y)\in\{0,1\}$ indicate whether a complete solution $y$ is correct. We define teacher recoverability as its expected accuracy when continuing from $h_t$,
$\Lambda^T(h_t)=\mathbb{E}_{y\sim\pi_T(\cdot\mid h_t)}[r(y)]$,
where $y$ denotes the completed solution. Student recoverability $\Lambda^S(h_t)$ is defined analogously using $\pi_\theta$.

$\Lambda^T$ measures the reliability of the teacher's continuation policy as a corrective target. Under the sampled reverse-KL objective, a student-sampled token $y_t$ receives the advantage
\setlength{\abovedisplayskip}{3pt}\setlength{\belowdisplayskip}{3pt}%
\begin{equation}
A_t = \log \pi_T(y_t \mid h_t)
    - \log \pi_\theta(y_t \mid h_t),
\label{eq:adv}
\end{equation}
which is positive when the teacher assigns $y_t$ a higher probability than the student. The OPD target at each prefix is therefore $\pi_T(\cdot\mid h_t)$. A student that perfectly imitates the teacher from $h_t$ onward would achieve accuracy $\Lambda^T(h_t)$. Thus, low recoverability indicates that the teacher's continuation policy is an unreliable corrective target, even though individual tokens may still be useful.

We measure both quantities using a Qwen3-1.7B student after 80 steps of OPD, the Qwen3-4B-Instruct-2507 teacher, and DAPO-Math-17k problems. The student generates one rollout for each of 900 problems, from which we randomly select 150 correct and 150 incorrect rollouts. We cut each rollout at the empty prefix and the seven paragraph boundaries nearest to $1/8,\ldots,7/8$ of its length. From each cut, the teacher and student each generate 16 continuations to estimate $\Lambda^T$ and $\Lambda^S$.

\begin{figure}[t]
\centering
\includegraphics[width=\columnwidth,trim=0 7pt 0 0,clip]{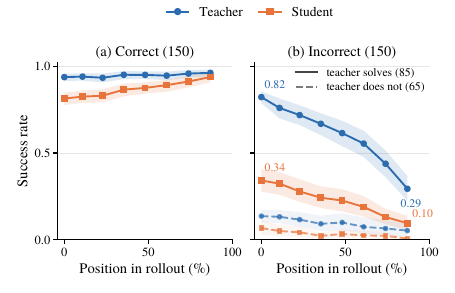}
\par\vspace{-4pt}
\caption{Teacher ($\Lambda^T$) and student ($\Lambda^S$) continuation accuracy along (a) correct and (b) incorrect rollouts.}
\label{fig:failure-region}
\vspace{6pt}
\includegraphics[width=\columnwidth]{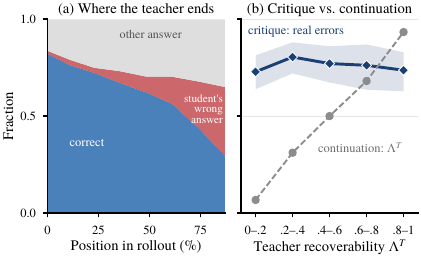}
\par\vspace{-4pt}
\caption{(a) Outcomes of teacher continuations. (b) Valid error-flag rate across teacher-recoverability bins.}
\label{fig:critique}
\end{figure}

To distinguish prefix effects from problem difficulty, we divide incorrect rollouts by the success rate of the teacher on the original problem, which is $\Lambda^T$ at the empty prefix estimated from $16$ continuations: 85 come from problems with high initial teacher success ($\Lambda^T \geq 0.5$; solid lines in Figure~\ref{fig:failure-region}(b)), and 65 from problems with low initial success (dashed lines). Appendix~\ref{app:diagnosis} gives details and repeats the analysis with Qwen3-0.6B.

\subsection{Recoverability Declines Along Incorrect Rollouts}
\label{sec:failure-region}

\noindent\textbf{The teacher retains recoverability on correct rollouts.}
On correct rollouts (Figure~\ref{fig:failure-region}(a)), $\Lambda^T$ stays at or above 0.93, while $\Lambda^S$ rises from 0.81 to 0.94. Thus, prefix growth alone does not explain the decline in teacher accuracy.

\noindent\textbf{Incorrect rollouts reduce teacher recoverability.}
On problems with high initial teacher success (solid lines in Figure~\ref{fig:failure-region}(b)), $\Lambda^T$ falls from 0.82 to 0.29, and $\Lambda^S$ from 0.34 to 0.10, consistent with prior observations \citep{liu2026sfd, zhou2026early, li2026rethinking}. Together with the control in Figure~\ref{fig:failure-region}(a), this suggests that the decline is associated with erroneous student prefixes rather than merely their length or problem difficulty. On problems with low initial teacher success (dashed lines), $\Lambda^T$ falls from 0.14 to 0.05, indicating that limited teacher capability is another source of unreliable supervision. Thus, OPD faces two difficulties: erroneous student prefixes can mislead an otherwise capable teacher, while some problems exceed the teacher's own ability.

\noindent\textbf{More than half of the tokens in incorrect rollouts receive supervision at low-recoverability prefixes.}
We assign each token to its preceding cut point and group tokens by $\Lambda^T$ at that cut. Among tokens in incorrect rollouts, 53\% correspond to prefixes with $\Lambda^T<0.2$, while only 21\% correspond to $\Lambda^T\geq0.8$. Since incorrect rollouts account for 61\% of training tokens, approximately one third of all training tokens receive supervision at prefixes from which the teacher rarely continues correctly.

\subsection{The Teacher Follows the Error but Still Recognizes It}
\label{sec:critique-ability}

We next examine why the teacher loses recoverability and whether it can still recognize errors. Both tests use the 85 incorrect rollouts from problems with high initial teacher success in Section~\ref{sec:failure-region}, without modifying the original trajectories.

First, we examine the final answers of teacher continuations from each cut. Second, we divide each nonempty prefix into numbered paragraphs and ask the teacher to identify its errors using our critique prompt (Appendix~\ref{app:prompts}). Unlike the critic in Section~\ref{sec:method}, the teacher receives no reference answer in this test. Qwen3-8B in thinking mode then judges each flagged error with access to the reference answer, labeling it as a new error, a propagated error, a restatement of an earlier error, or not an error. We count the first three categories as valid (Appendix~\ref{app:diagnosis}). This measures the validity of flagged errors rather than error-detection recall.

\noindent\textbf{The teacher often inherits the student's erroneous conclusions.}
As the prefix grows longer (Figure~\ref{fig:critique}(a)), the fraction of correct teacher continuations falls from 0.82 to 0.29, while the fraction ending with the student's incorrect answer rises from 1\% to 35\%, or from 8\% to 50\% of failed continuations. Since the teacher rarely produces this answer from the empty prefix, these results suggest that it often follows errors in the student prefix rather than failing solely because it cannot solve the original problem.

\noindent\textbf{Error recognition precision remains stable as recoverability declines.}
Across five bins of $\Lambda^T$ (Figure~\ref{fig:critique}(b)), mean recoverability ranges from 0.07 to 0.93, while the fraction of valid error flags remains between 73\% and 81\%. This fraction is 73\% at $\Lambda^T<0.2$ and 74\% at $\Lambda^T\geq0.8$. Thus, teacher continuation accuracy declines sharply, but critique precision remains stable.

\section{CriticOPD: Restoring Recoverability with Critique}
\label{sec:method}

Section~\ref{sec:diagnosis} shows that after the student errs, the teacher may fail to continue correctly but can still recognize the errors. We therefore build training sequences that the student writes from the states that it reaches at inference, which adds little occupancy mismatch, and on which the teacher supervision is reliable. Section~\ref{sec:rewrite} builds these sequences and measures the recoverability of the teacher on them, and Section~\ref{sec:fbd} gives the training objective.

\subsection{Critique and Rewrite}
\label{sec:rewrite}

\noindent\textbf{Trigger.}
The student generates a rollout $y$ for each problem $x$ with the prompt $\mathcal{P}_{\text{train}}$. Incorrect rollouts and those reaching the length limit without an answer are sent for critique and rewriting. Correct rollouts are trained with standard OPD.

\noindent\textbf{Critique.}
The teacher serves as the critic. It receives the problem, the reference answer, and the solution of the student in numbered paragraphs, and it lists the errors in the solution (prompts $\mathcal{P}_{\text{critic}}$ and $\mathcal{P}_{\text{critic-user}}$ in Appendix~\ref{app:prompts-method}). Each error contains the index of its paragraph, the erroneous sentence, and a short fix. We combine these errors and the reference answer into the critique $c$. The reference answer lets the teacher critique problems that it cannot solve. If the critique contains no complete error, the rollout is trained with standard OPD.

\noindent\textbf{Cut point.}
The student keeps its prefix through the paragraph of the last error in the critique and discards the rest. The discarded part starts at position $\tau$, so the retained prefix is $h_\tau=(x,y_{<\tau})$. This prefix contains all errors in the critique, so the rewrite starts from the erroneous states where teacher recoverability declines (Section~\ref{sec:diagnosis}).

\begin{figure*}[t]
\centering
\includegraphics[width=\textwidth]{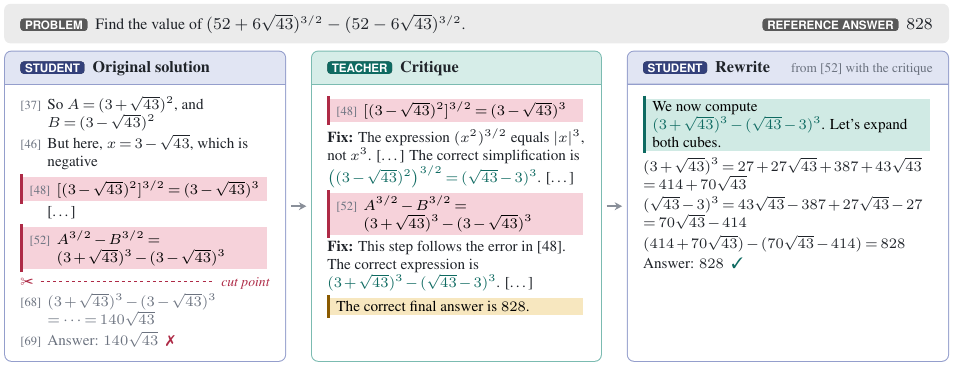}
\caption{An example of critique-guided rewriting. Left: the original solution, with the critiqued paragraphs marked. Middle: the critique, which quotes each erroneous sentence, gives a fix, and ends with the reference answer. Right: the suffix that the student writes from the last error with the critique.}
\label{fig:example}
\end{figure*}

\noindent\textbf{Rewrite.}
The student appends the critique $c$ to the retained prefix with the prompt $\mathcal{P}_{\text{fb}}$ and generates a new suffix, which gives the revised rollout $\tilde{y}$. The tokens before $\tau$ are unchanged, $\tilde{y}_{<\tau}=y_{<\tau}$, and the student samples the suffix with the critique in its context. Unlike teacher takeover, the student writes every token of the training sequence. In Figure~\ref{fig:example}, the critique identifies two errors, and the student rewrites from the last one to reach the correct answer.

\begin{figure}[t]
\centering
\includegraphics[width=\columnwidth]{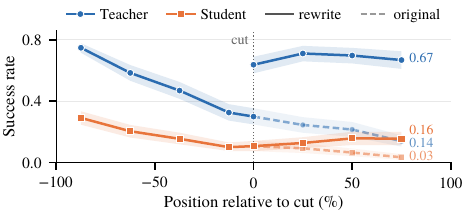}
\caption{Recoverability of the teacher and the student on original rollouts (dashed) and critique-guided rewrites (solid), plotted against the position relative to the cut point. On the rewrite, the teacher sees the critique, and the student does not.}
\label{fig:rewrite-effect}
\end{figure}

\noindent\textbf{Recoverability on the rewrite.}
We test whether the teacher is more reliable on the rewrite than on the original rollout. As in Section~\ref{sec:diagnosis}, we estimate recoverability as the fraction of correct answers among eight continuations from each prefix. The test uses the student after 80 steps of OPD and the 161 incorrect rollouts of Appendix~\ref{app:rewrite-analysis}. We critique and cut each rollout as CriticOPD does, and the student writes a rewrite with the critique. This student is not trained with CriticOPD, so the rewrite differs from the original rollout only through the critique. Figure~\ref{fig:rewrite-effect} compares recoverability after the cut point on the two sequences. On the original rollout, the teacher continues from the prefix of the student, and on the rewrite, the teacher continues with the critique in its context. At 75\% of the continuation, the recoverability of the teacher is 0.14 on the original rollout and 0.67 on the rewrite, and the success rate of the student without the critique rises from 0.03 to 0.16.

\subsection{Training Objective}
\label{sec:fbd}
\label{sec:privileged}

\noindent\textbf{Standard OPD.}
For correct rollouts and rollouts without a valid critique, we minimize the reverse KL $\mathcal{L}_{\mathrm{OPD}}(y)=\sum_{t=1}^{|y|}\mathrm{KL}\big(\pi_\theta(\cdot\mid h_t)\,\|\,\pi_T(\cdot\mid h_t)\big)$ at each student-generated prefix $h_t$ with the sampled-token advantage in Equation~\eqref{eq:adv}.

\noindent\textbf{Feedback distillation.}
For rollouts with a valid critique, we train on the revised sequence $\tilde{y}$. Let $\tilde{h}_t=(x,\tilde{y}_{<t})$ denote its prefix without the critique. The teacher scores the rewrite with $c$ in its context, while the student does not see it:
\begin{equation}
\mathcal{L}_{\mathrm{fb}}(\tilde{y};c)
=
\sum_{t=\tau}^{|\tilde{y}|}
\mathrm{KL}\left(
\pi_\theta(\cdot\mid\tilde{h}_t)
\,\big\|\,
\pi_T(\cdot\mid\tilde{h}_t,c)
\right).
\label{eq:fb-loss}
\end{equation}
Only the rewritten suffix receives supervision, since the retained prefix contains errors at states where the teacher has low recoverability. As in OPD, each suffix token receives the sampled-token advantage
\begin{equation}
A_t^{\mathrm{fb}}
=
\log\pi_T(\tilde{y}_t\mid\tilde{h}_t,c)-\log\pi_\theta(\tilde{y}_t\mid\tilde{h}_t).
\label{eq:fb}
\end{equation}
The suffix is sampled with the critique, whereas the log-probabilities of the student are computed without it, so this update is a surrogate rather than an unbiased gradient of Equation~\eqref{eq:fb-loss}.

\noindent\textbf{Overall objective.}
CriticOPD minimizes $\mathcal{L}_{\mathrm{ours}}=\mathcal{L}_{\mathrm{OPD}}+\mathcal{L}_{\mathrm{fb}}$, where each loss sums over its rollouts in the batch and each rollout enters exactly one term. Compared with OPD, the teacher target changes from $\pi_T(\cdot\mid h_t)$ to $\pi_T(\cdot\mid\tilde{h}_t,c)$, under which the teacher is much more recoverable (Section~\ref{sec:rewrite}). The student learns without the critique and needs no feedback at inference. Algorithm~\ref{alg:criticopd} in Appendix~\ref{app:config} summarizes one training step.

\section{Experiments}
\label{sec:experiments}

The experiments answer three questions.
\textbf{RQ1}: does critique-guided rewriting improve on-policy distillation on
held-out mathematical reasoning benchmarks, and does it preserve performance
outside mathematics?
\textbf{RQ2}: which components of \ours{} account for the improvement?
\textbf{RQ3}: where do the gains of \ours{} come from?
Section~\ref{sec:experiments-setup} describes the shared setup, and the
following subsections answer each question in turn.

\subsection{Setup}
\label{sec:experiments-setup}

\noindent\textbf{Models and training.}
We distill Qwen3-1.7B and Qwen3-0.6B, both in non-thinking mode, from the
Qwen3-4B-Instruct-2507 teacher on the prompts of DAPO-Math-17k.
All methods within a student scale share the same training recipe.
\ours{} uses the teacher as the critic, and the critic lists all errors that it
finds in the solution of the student within a limit of $2{,}048$ tokens.
Appendix~\ref{app:setup} gives the hyperparameters, and
Appendix~\ref{app:prompts} lists the prompts.

\noindent\textbf{Baselines.}
We compare four classes of baselines.
The first class contains standard post-training methods.
SFT and KD train on solutions of the teacher, GRPO uses an outcome
reward without the teacher, and OPD performs standard on-policy distillation.

The second class uses privileged information during
training, as CriticOPD does. OPSD-T
\citep{zhao2026selfdistilled} lets the teacher score
student rollouts with the reference answer in its
context. Critique-GRPO \citep{zhang2026critiquegrpo}
lets the student revise incorrect solutions with
teacher critiques and optimizes the original and
revised solutions with outcome rewards.
The third and fourth classes correspond to the two responses to the loss of
recoverability that Section~\ref{sec:intro} describes.
The third class reduces supervision, and FastOPD \citep{zhang2026fast}
represents it by truncating long rollouts.
The fourth class replaces the sequence, and TRD \citep{jiang2026trajectory},
SKD \citep{xu2025speculative}, and RelayOPD \citep{xu2026relay} let the teacher
rewrite the solution of the student or write into its trajectory.

\noindent\textbf{Evaluation.}
We evaluate on AIME 2024, AIME 2025, AMC 2023, MATH500, OlympiadBench, and
Minerva Math with a budget of $32{,}768$ tokens.
We report the mean accuracy over $32$ samples per problem on the first three
benchmarks and over $8$ samples on the others.
We train each method with three seeds and report the mean and the standard
deviation at the checkpoint that we select under seed $42$
(Appendix~\ref{app:setup}).

\subsection{RQ1: Does critique-guided rewriting improve OPD?}
\label{sec:experiments-results}

\begin{table*}[t]
\centering
\caption{
\textbf{Results on six mathematical reasoning benchmarks.}
Entries are means and standard deviations over three seeds, and the subscript after each method gives the selected step.
Bold marks the best student at each scale.
}
\label{tab:main}

\scriptsize
\setlength{\tabcolsep}{3pt}
\renewcommand{\arraystretch}{1.05}

\begin{tabularx}{\textwidth}{l*{7}{>{\centering\arraybackslash}X}}
\toprule
\textbf{Method}
& \textbf{AIME24}
& \textbf{AIME25}
& \textbf{AMC23}
& \textbf{MATH500}
& \textbf{OlympiadBench}
& \textbf{Minerva}
& \textbf{Avg}
\\
\midrule

Teacher
& 62.19 & 48.12 & 85.99 & 94.83 & 72.26 & 45.54
& 68.16 \\

\midrule
\multicolumn{8}{c}{\textit{Student: Qwen3-0.6B-Non-Thinking}} \\
\midrule

Student
& 2.08 & 1.35 & 18.98 & 45.52 & 16.94 & 13.69
& 16.43 \\

\arrayrulecolor{black!35}\specialrule{0.35pt}{1.2pt}{1.2pt}\arrayrulecolor{black}

SFT$_{@139}$
& 4.62$_{\pm 0.49}$ & 6.84$_{\pm 0.12}$ & 30.08$_{\pm 0.20}$ & 58.60$_{\pm 0.37}$ & 27.24$_{\pm 0.20}$ & 15.59$_{\pm 0.25}$
& 23.83$_{\pm 0.06}$ \\

KD$_{@139}$
& 4.51$_{\pm 0.63}$ & 6.15$_{\pm 0.73}$ & 28.75$_{\pm 0.80}$ & 57.88$_{\pm 0.69}$ & 26.62$_{\pm 0.52}$ & 15.09$_{\pm 0.88}$
& 23.17$_{\pm 0.49}$ \\

GRPO$_{@80}$
& 3.82$_{\pm 0.71}$ & 3.78$_{\pm 1.27}$ & 26.46$_{\pm 0.92}$ & 53.72$_{\pm 1.60}$ & 23.45$_{\pm 1.14}$ & 15.12$_{\pm 0.71}$
& 21.06$_{\pm 1.00}$ \\

Critique-GRPO$_{@40}$
& 1.42$_{\pm 0.43}$ & 0.97$_{\pm 0.53}$ & 16.59$_{\pm 0.61}$ & 41.02$_{\pm 1.07}$ & 13.93$_{\pm 0.45}$ & 11.47$_{\pm 0.15}$
& 14.23$_{\pm 0.49}$ \\

\arrayrulecolor{black!35}\specialrule{0.35pt}{1.2pt}{1.2pt}\arrayrulecolor{black}

OPD$_{@40}$
& 10.24$_{\pm 1.17}$ & 14.69$_{\pm 1.36}$ & 45.14$_{\pm 0.64}$ & 72.93$_{\pm 0.32}$ & 42.42$_{\pm 0.79}$ & 23.35$_{\pm 0.20}$
& 34.80$_{\pm 0.30}$ \\

OPSD-T$_{@80}$
& 13.37$_{\pm 0.93}$ & 15.00$_{\pm 0.28}$ & 47.48$_{\pm 0.25}$ & 75.52$_{\pm 0.43}$ & 44.10$_{\pm 0.26}$ & 24.26$_{\pm 0.44}$
& 36.62$_{\pm 0.15}$ \\

\arrayrulecolor{black!35}\specialrule{0.35pt}{1.2pt}{1.2pt}\arrayrulecolor{black}

FastOPD$_{@40}$
& 13.72$_{\pm 0.95}$ & 18.75$_{\pm 0.91}$ & 47.43$_{\pm 0.94}$ & 75.18$_{\pm 0.29}$ & 44.48$_{\pm 0.65}$ & 24.17$_{\pm 0.63}$
& 37.29$_{\pm 0.46}$ \\

TRD$_{@139}$
& 3.75$_{\pm 0.31}$ & 5.35$_{\pm 0.37}$ & 27.85$_{\pm 0.51}$ & 55.14$_{\pm 0.55}$ & 25.72$_{\pm 0.60}$ & 14.86$_{\pm 0.25}$
& 22.11$_{\pm 0.25}$ \\

SKD$_{@40}$
& 7.78$_{\pm 0.63}$ & 10.38$_{\pm 0.53}$ & 36.19$_{\pm 0.45}$ & 65.60$_{\pm 0.21}$ & 33.81$_{\pm 0.53}$ & 19.41$_{\pm 0.56}$
& 28.86$_{\pm 0.24}$ \\

RelayOPD$_{@60}$
& 13.90$_{\pm 0.58}$ & 17.64$_{\pm 0.57}$ & 48.96$_{\pm 1.11}$ & 76.49$_{\pm 0.31}$ & 46.01$_{\pm 0.25}$ & \textbf{25.60}$_{\pm 0.44}$
& 38.10$_{\pm 0.22}$ \\

\rowcolor{blue!8}
\textbf{\ours{}}$_{@40}$
& \textbf{14.81}$_{\pm 0.73}$ & \textbf{19.41}$_{\pm 0.47}$ & \textbf{50.44}$_{\pm 0.48}$ & \textbf{76.56}$_{\pm 0.35}$ & \textbf{46.81}$_{\pm 0.51}$ & 25.14$_{\pm 0.26}$
& \textbf{38.86}$_{\pm 0.15}$ \\

\rowcolor{blue!8}
\textcolor{green!50!black}{$\Delta$ vs OPD}
& \textcolor{green!50!black}{+4.57}
& \textcolor{green!50!black}{+4.72}
& \textcolor{green!50!black}{+5.30}
& \textcolor{green!50!black}{+3.63}
& \textcolor{green!50!black}{+4.39}
& \textcolor{green!50!black}{+1.79}
& \textcolor{green!50!black}{+4.06}
\\

\midrule
\multicolumn{8}{c}{\textit{Student: Qwen3-1.7B-Non-Thinking}} \\
\midrule

Student
& 11.98 & 9.17 & 41.00 & 73.22 & 39.80 & 29.23
& 34.07 \\

\arrayrulecolor{black!35}\specialrule{0.35pt}{1.2pt}{1.2pt}\arrayrulecolor{black}

SFT$_{@40}$
& 27.71$_{\pm 1.00}$ & 23.37$_{\pm 1.03}$ & 58.82$_{\pm 0.40}$ & 83.39$_{\pm 0.38}$ & 52.58$_{\pm 0.42}$ & 34.56$_{\pm 0.09}$
& 46.74$_{\pm 0.48}$ \\

KD$_{@139}$
& 20.59$_{\pm 0.43}$ & 17.85$_{\pm 2.29}$ & 52.18$_{\pm 0.63}$ & 79.22$_{\pm 0.22}$ & 48.50$_{\pm 0.46}$ & 33.79$_{\pm 0.92}$
& 42.02$_{\pm 0.49}$ \\

GRPO$_{@120}$
& 26.94$_{\pm 1.59}$ & 20.80$_{\pm 1.20}$ & 56.22$_{\pm 0.43}$ & 82.21$_{\pm 0.98}$ & 52.15$_{\pm 1.04}$ & 33.95$_{\pm 0.53}$
& 45.38$_{\pm 0.85}$ \\

Critique-GRPO$_{@40}$
& 9.83$_{\pm 0.57}$ & 8.58$_{\pm 0.39}$ & 38.24$_{\pm 0.28}$ & 70.11$_{\pm 0.34}$ & 35.90$_{\pm 0.25}$ & 28.43$_{\pm 0.31}$
& 31.85$_{\pm 0.25}$ \\

\arrayrulecolor{black!35}\specialrule{0.35pt}{1.2pt}{1.2pt}\arrayrulecolor{black}

OPD$_{@80}$
& 35.10$_{\pm 0.42}$ & 22.78$_{\pm 0.42}$ & 62.94$_{\pm 0.40}$ & 86.26$_{\pm 0.42}$ & 56.65$_{\pm 0.36}$ & 36.78$_{\pm 0.53}$
& 50.08$_{\pm 0.09}$ \\

OPSD-T$_{@60}$
& 37.08$_{\pm 0.73}$ & 23.51$_{\pm 0.51}$ & 63.64$_{\pm 0.30}$ & 86.85$_{\pm 0.33}$ & 57.47$_{\pm 0.19}$ & \textbf{37.53}$_{\pm 0.79}$
& 51.01$_{\pm 0.39}$ \\

\arrayrulecolor{black!35}\specialrule{0.35pt}{1.2pt}{1.2pt}\arrayrulecolor{black}

FastOPD$_{@60}$
& 40.97$_{\pm 1.84}$ & 26.25$_{\pm 0.58}$ & 66.08$_{\pm 0.24}$ & 87.28$_{\pm 0.36}$ & 58.01$_{\pm 0.27}$ & 36.73$_{\pm 0.99}$
& 52.55$_{\pm 0.36}$ \\

TRD$_{@40}$
& 21.15$_{\pm 0.54}$ & 18.72$_{\pm 0.57}$ & 50.95$_{\pm 0.85}$ & 79.31$_{\pm 0.45}$ & 46.38$_{\pm 0.49}$ & 32.05$_{\pm 0.10}$
& 41.42$_{\pm 0.15}$ \\

SKD$_{@120}$
& 26.42$_{\pm 1.00}$ & 21.53$_{\pm 1.22}$ & 56.84$_{\pm 0.13}$ & 83.00$_{\pm 0.55}$ & 52.07$_{\pm 0.47}$ & 36.63$_{\pm 0.42}$
& 46.08$_{\pm 0.54}$ \\

RelayOPD$_{@40}$
& 40.10$_{\pm 1.15}$ & 29.97$_{\pm 0.60}$ & 67.32$_{\pm 0.52}$ & 88.10$_{\pm 0.35}$ & 59.07$_{\pm 0.25}$ & 37.50$_{\pm 0.12}$
& 53.68$_{\pm 0.39}$ \\

\rowcolor{blue!8}
\textbf{\ours{}}$_{@40}$
& \textbf{41.11}$_{\pm 1.02}$ & \textbf{30.14}$_{\pm 1.00}$ & \textbf{69.88}$_{\pm 0.26}$ & \textbf{88.87}$_{\pm 0.50}$ & \textbf{60.45}$_{\pm 0.51}$ & 37.44$_{\pm 0.19}$
& \textbf{54.65}$_{\pm 0.10}$ \\

\rowcolor{blue!8}
\textcolor{green!50!black}{$\Delta$ vs OPD}
& \textcolor{green!50!black}{+6.01}
& \textcolor{green!50!black}{+7.36}
& \textcolor{green!50!black}{+6.94}
& \textcolor{green!50!black}{+2.61}
& \textcolor{green!50!black}{+3.80}
& \textcolor{green!50!black}{+0.66}
& \textcolor{green!50!black}{+4.57}
\\

\bottomrule
\end{tabularx}
\end{table*}


Table~\ref{tab:main} reports the main results.
\ours{} improves OPD by $4.57$ points ($50.08\rightarrow54.65$)
on Qwen3-1.7B and $4.06$ points ($34.80\rightarrow38.86$)
on Qwen3-0.6B.
On Qwen3-1.7B, the largest gains occur on AIME 2024, AIME 2025,
and AMC 2023 ($+6.01$, $+7.36$, and $+6.94$).
\ours{} achieves the best student accuracy on five of six
benchmarks at both scales.
On Qwen3-1.7B, it outperforms FastOPD and RelayOPD by $2.10$
and $0.97$ points, respectively.
On Qwen3-0.6B, the corresponding gains are $1.57$ and $0.76$ points.

Among baselines using privileged information, OPSD-T yields
smaller gains over OPD ($+0.93$ and $+1.82$ at the two scales),
while Critique-GRPO underperforms the initial students.
These results suggest that answer-conditioned scoring or
critique-guided revision alone is insufficient to match \ours{}.
The improvement also extends to OLMo-3 models, where \ours{}
raises the average from $54.69$ to $57.23$
(Appendix~\ref{app:olmo}).
Outside mathematics, \ours{} achieves an average of $70.69$,
comparable to OPD ($70.47$), with modest declines on code
benchmarks (Appendix~\ref{app:ood}).
Training dynamics and computational costs
are provided in Appendices~\ref{app:results} and~\ref{app:cost}.

\subsection{RQ2: Which components of \ours{} matter?}
\label{sec:experiments-ablation}

\begin{table}[t]
\centering
\caption{\textbf{Ablations} on Qwen3-1.7B with seed $42$. $\Delta$ is the
change in Avg from \ours{}, and from \ours{} without the answer in (b) and (c).
AIME averages AIME 2024 and AIME 2025.}
\label{tab:ablation}
\footnotesize
\setlength{\tabcolsep}{3.5pt}
\renewcommand{\arraystretch}{1.06}
\newcommand{\up}[1]{\textcolor{green!45!black}{$+$#1}}
\newcommand{\dn}[1]{\textcolor{red!65!black}{$-$#1}}
\newcommand{\blk}[1]{\multicolumn{4}{@{}l}{\textit{#1}}}
\newcommand{\tbd}{\textcolor{red}{TBD}}
\begin{tabular}{@{}lccr@{}}
\toprule
\textbf{Variant} & \textbf{AIME} & \textbf{Avg} & $\boldsymbol{\Delta}$ \\
\midrule
\rowcolor{blue!8}
\ours{} & \textbf{34.79} & \textbf{54.57} & -- \\
\midrule
\blk{(a) Content of the feedback} \\
\quad \textit{w/o} answer & 34.58 & 53.72 & \dn{0.85} \\
\quad \textit{w/o} critique & 28.22 & 48.67 & \dn{5.90} \\
\quad critique + answer $\rightarrow$ generic & 33.85 & 53.11 & \dn{1.46} \\
\midrule
\blk{(b) Writer of the suffix} \\
\quad student $\rightarrow$ teacher & 33.18 & 53.06 & \dn{0.66} \\
\midrule
\blk{(c) Scoring of the suffix} \\
\quad \textit{w/o} feedback & 33.02 & 52.53 & \dn{1.19} \\
\midrule
OPD & 28.70 & 50.18 & \dn{4.39} \\
\bottomrule
\end{tabular}
\end{table}

Table~\ref{tab:ablation} changes one component of \ours{} at a time.
The components correspond to three designs of Section~\ref{sec:method}: the
feedback that the student sees during the rewrite, the model that writes the
suffix after the cut point, and whether the teacher sees the feedback when it
scores the suffix.
All variants use the Qwen3-1.7B student with seed $42$, so differences below
about $0.5$ points are comparable to the standard deviation across seeds in
Table~\ref{tab:main}.
Appendix~\ref{app:ablation} reports all six benchmarks for each variant, and
Appendix~\ref{app:hyper} varies the length limit of the critique.

\noindent\textbf{Content of the feedback.}
The feedback of \ours{} contains the critique and the reference answer.
Removing the reference answer lowers the average by $0.85$, removing the
critique lowers it by $5.90$, and replacing both with generic feedback lowers
it by $1.46$.
With the answer alone, the student falls even below OPD, likely because the
rewrite then reaches the given answer without a corrected derivation, so the
student learns to state answers rather than the reasoning that leads to them.

\noindent\textbf{Writer of the suffix.}
In this variant, the teacher continues from the same cut point with the same
feedback, and the student is trained on the suffix of the teacher with the
same loss.
This variant lowers the average by $0.66$.

\noindent\textbf{Scoring of the suffix.}
When the teacher scores the suffix of the student without the feedback, the
average falls by $1.19$.
In this case, the teacher scores the correction of the student at states after
its error, and Section~\ref{sec:diagnosis} shows that the recoverability of the
teacher is low at these states.
The critique therefore improves both the suffix that the student writes and the
reliability of the teacher that scores it, and
Section~\ref{sec:experiments-why} analyzes this effect.
\subsection{RQ3: Where do the gains of \ours{} come from?}
\label{sec:experiments-why}

Section~\ref{sec:rewrite} shows that the critique substantially raises the recoverability of the teacher on the rewrite.
This section answers two questions: whether the supervision of \ours{} meets both requirements on reliability and occupancy mismatch, and how the more reliable target of the teacher becomes an ability of the student.

\begin{figure}[t]
\centering
\includegraphics[width=\columnwidth]{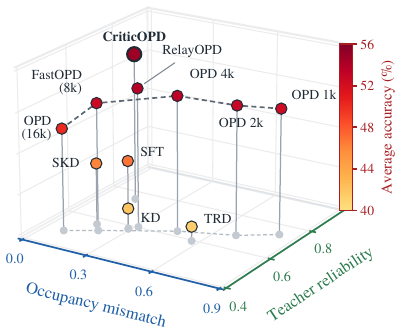}
\caption{Occupancy mismatch and teacher reliability on the states that each method supervises, and the average accuracy of the Qwen3-1.7B student.}
\label{fig:mismatch-reliability}
\end{figure}

\noindent\textbf{Reliability and occupancy mismatch together determine accuracy.}
Figure~\ref{fig:mismatch-reliability} places each method by the occupancy mismatch and the teacher reliability on the states that it supervises, and it marks the average accuracy of the Qwen3-1.7B student.
The occupancy mismatch is the total variation distance between the token positions of the prefixes that a method supervises and those of the prefixes that the student reaches at inference.
This distance is a lower bound on the true mismatch, and it shows which part of the response the supervision covers.
The teacher reliability is the mean recoverability of the teacher on the supervised states (Appendix~\ref{app:occupancy}).
\ours{} has an occupancy mismatch of $0.16$, which is slightly above $0.10$ for OPD but below $0.21$ for FastOPD and $0.35$ for RelayOPD.
FastOPD does not supervise positions after $8{,}192$ tokens, and RelayOPD ends the rollout after the second takeover, so both remove the later part of the response that the student reaches at inference.
The rewrite of \ours{} instead runs from the cut point to the end of the response.
A shorter length limit for OPD also increases the occupancy mismatch, and the accuracy stops improving beyond 8k tokens.
\ours{} combines the highest teacher reliability with a low occupancy mismatch and reaches the highest average accuracy.

\begin{figure}[t]
\centering
\includegraphics[width=\columnwidth]{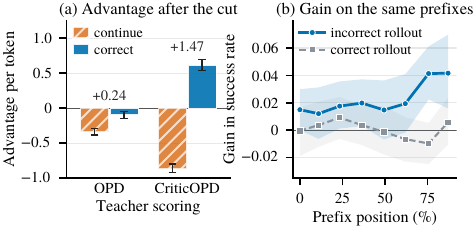}
\caption{(a) Advantage per token of two continuations of $64$ tokens at the cut point, which either continue or correct the error, under the scoring of OPD and of \ours{}. Numbers give correct minus continue. (b) Success rate of the student of \ours{} minus that of OPD from the same prefixes on OlympiadBench, with 95\% confidence bands.}
\label{fig:why}
\end{figure}

\noindent\textbf{The critique gives credit to the correction.}
For each token of the rewrite, the advantage in Equation~\eqref{eq:fb} decomposes exactly as
\begin{equation}
\begin{aligned}
A_t^{\mathrm{fb}}
={}&
\underbrace{\log\pi_T(\tilde{y}_t\mid\tilde{h}_t)-\log\pi_\theta(\tilde{y}_t\mid\tilde{h}_t)}_{\text{OPD advantage}}\\
&+
\underbrace{\log\frac{\pi_T(\tilde{y}_t\mid\tilde{h}_t,c)}{\pi_T(\tilde{y}_t\mid\tilde{h}_t)}}_{r_t}.
\end{aligned}
\label{eq:fb-decomp}
\end{equation}
The first term is the advantage of OPD on the same sequence, and the second term $r_t$ is the pointwise mutual information between the token and the critique under the teacher.
Feedback distillation is therefore OPD on the rewrite with an additional reward $r_t$ on each token, and through this reward the teacher turns its recognition of errors into token-level credit.
To measure this credit, we compare two continuations of $64$ tokens at $132$ cut points, of which one continues the error and the other corrects it.
Under the scoring of OPD, the correction leads by only $0.24$ per token (95\% CI $[0.19, 0.29]$), whereas under the scoring of \ours{}, it leads by $1.47$ ($[1.36, 1.58]$; Figure~\ref{fig:why}(a)).
The first $32$ tokens account for $87\%$ of this increase.
A teacher without the critique thus barely separates the correction from the error, which is consistent with the drop of 1.19 points when the teacher scores the suffix without the feedback (Section~\ref{sec:experiments-ablation}).
The credit from the critique does not depend on which model writes the suffix, so a suffix that the teacher writes receives similar credit, consistent with the smaller drop of 0.66 points in that case.

\noindent\textbf{The student learns to recover from its own errors.}
We take $150$ incorrect and $150$ correct rollouts of the initial student on OlympiadBench, which is not part of the training data, and let the students of OPD and \ours{} continue $16$ times from the empty prefix and each cut point.
The two students differ by at most $0.015$ on the empty prefix and by $0.000$ on average on correct prefixes.
On incorrect prefixes, the student of \ours{} leads by $0.023$ on average ($[0.010, 0.036]$) and by about $0.04$ at the last two cut points (Figure~\ref{fig:why}(b)).
The gain therefore concentrates on the states after the student errs, which are the states where the supervision of OPD loses its reliability in Section~\ref{sec:diagnosis}.
This behavior also lengthens the responses, so part of the gain comes with longer reasoning, but at the same response length the critique still improves the student (Appendix~\ref{app:budget}).
\section{Related Work}
\label{sec:related}

\noindent\textbf{On-policy distillation and its supervision.}
On-policy distillation trains the student on its own sequences with feedback from the teacher \citep{agarwal2024onpolicy,gu2023minillm,ko2024distillm,yang2026eos}, and its transfer depends on whether the reasoning of the student and the teacher is compatible \citep{li2026rethinking}.
As prefixes degrade, the feedback of the teacher can lose its corrective value \citep{liu2026prefix,xin2026escaping}, and existing methods truncate, select, or reweight the supervision \citep{zhang2026fast,yang2026pruneopd,wang2026teachability,zheng2026scope,lee2026sead}.
These methods choose where to supervise, whereas \ours{} rewrites the rollout after its last error.

\noindent\textbf{Teacher intervention and trajectory repair.}
Other methods let the teacher write tokens into the trajectory through interleaved generation, revisions, or takeovers \citep{xu2025speculative,jiang2026trajectory,xu2026relay}, or regenerate from an earlier reliable context \citep{wang2026backtracking}.
In \ours{}, the teacher writes only a critique, which is removed before training, so every supervised token comes from the student.

\noindent\textbf{Privileged context and language feedback.}
Context distillation trains a model to match itself with a fixed context \citep{askell2021general,snell2022learning}, and OPSD applies it to OPD with the reference answer as the context \citep{zhao2026selfdistilled,yang2026ogls}.
Other work trains on revisions guided by language feedback \citep{scheurer2023training,zhang2026critiquegrpo} or on self-correction in a later turn \citep{kumar2025training,qu2024recursive}.
In \ours{}, the teacher writes the critique for each incorrect rollout, and the student needs neither feedback nor an extra turn at inference.

\section{Conclusion}
\label{sec:conclusion}

We show that on-policy distillation relies on the recoverability of the teacher, which fails after the student errs because the teacher follows the errors of the student while still recognizing them. CriticOPD uses the critique of the teacher to restore recoverability and improves OPD on mathematical reasoning benchmarks at two student scales. The critique changes only the context in which the teacher scores the rewrite, so the student writes every supervised token and needs no feedback at inference. It restores reliable token-level credit.

\section*{Limitations}

\noindent\textbf{Reference answers.}
CriticOPD uses reference answers to decide which rollouts receive a critique and
to ground the critic, so it applies directly to tasks with verifiable answers,
such as mathematics. Tasks without verifiable answers would need a reward model
or a critic that judges correctness on its own, which we leave to future work.
For code, unit tests could play the role of the reference answer.

\noindent\textbf{Models and domains.}
Our experiments distill Qwen3 students of two sizes from one Qwen3 teacher in
non-thinking mode, and an OLMo-3 student from an OLMo-3 teacher with one seed
(Appendix~\ref{app:olmo}), all on mathematical problems. The students preserve
their performance outside mathematics (Appendix~\ref{app:ood}), but students that
reason in thinking mode, training domains such as code, and larger teachers and
students remain to be tested.

\noindent\textbf{Training cost.}
Each critiqued rollout needs one more generation of the teacher and one more
generation of the student, although
CriticOPD reaches its selected step earlier and uses $0.42$ times the total
training compute of OPD (Appendix~\ref{app:cost}).

\noindent\textbf{Quality of the critique.}
The teacher serves as the critic, and $73\%$ to $81\%$ of the errors that it
lists are real under our judge (Section~\ref{sec:critique-ability}). CriticOPD
improves OPD with this critic, and a more accurate critic may further improve
the rewrite. We also use one critique per rollout, and sampling
several critiques could further reduce noise. The diagnosis also relies on a
model judge, which may disagree with human annotators in some cases.


%


\bibliography{custom}
\appendix
\clearpage
\setcounter{topnumber}{3}
\setcounter{dbltopnumber}{2}
\setcounter{totalnumber}{4}
\renewcommand{\topfraction}{0.9}
\renewcommand{\dbltopfraction}{0.9}
\renewcommand{\textfraction}{0.1}
\renewcommand{\floatpagefraction}{0.9}
\renewcommand{\dblfloatpagefraction}{0.9}

\section*{Appendix}

\startcontents[appendix]
\printcontents[appendix]{}{1}{\setcounter{tocdepth}{2}}

\vspace{1em}

\section{Prompts}
\label{app:prompts}

This appendix lists the prompts verbatim, with placeholders in typewriter font.
In training, the student receives each problem of DAPO-Math-17k in the user message $\mathcal{P}_{\text{train}}$, and in evaluation, each problem of the benchmarks in $\mathcal{P}_{\text{eval}}$.
In the other prompts, \texttt{\{student prompt\}} stands for the complete user message $\mathcal{P}_{\text{train}}$ of the problem.
The critic is the teacher model.
It receives the solution of the student split into paragraphs at blank lines, where headings, separator lines, and short lead-in lines are merged into the next paragraph, and the prompts call these paragraphs segments.

\begin{prompt}[title={$\mathcal{P}_{\text{eval}}$ --- user message of the student and the teacher in evaluation}]
\texttt{\{problem\}} Please reason step by step, and put your final answer within \verb|\boxed{}|.
\end{prompt}

\subsection{Prompts of \ours{}}
\label{app:prompts-method}

The critic of \ours{} receives the system prompt $\mathcal{P}_{\text{critic}}$, which defines the task and the output format, and the user message $\mathcal{P}_{\text{critic-user}}$, which gives the problem, the reference answer, and the numbered paragraphs of the solution.
For a rollout that reaches the length limit, the first paragraph of $\mathcal{P}_{\text{critic}}$ is replaced by $\mathcal{P}_{\text{critic}}^{\text{len}}$.
From the reply of the critic, we parse the listed errors and build the feedback, which ends with the reference answer.
\ours{} keeps the rollout up to the end of the paragraph of the last listed error and appends $\mathcal{P}_{\text{fb}}$, which contains the feedback.
The student continues after $\mathcal{P}_{\text{fb}}$, and the teacher scores this continuation with $\mathcal{P}_{\text{fb}}$ in its context, whereas the training sequence of the student does not contain $\mathcal{P}_{\text{fb}}$.

\begin{prompt}[title={$\mathcal{P}_{\text{train}}$ --- user message of the student in training (DAPO-Math-17k)}]
Solve the following math problem step by step. The last line of your response should be of the form Answer: \$Answer (without quotes) where \$Answer is the answer to the problem.\\[4pt]
\texttt{\{problem\}}\\[4pt]
Remember to put your answer on its own line after ``Answer:''.
\end{prompt}

\begin{prompt}[title={$\mathcal{P}_{\text{critic}}$ --- system prompt of the critic}]
You are a meticulous math teacher. A student has submitted a complete solution whose FINAL ANSWER IS WRONG. You are given the correct answer. Use it as a reference to trace the student's work forward and find the steps that contain genuine errors.\\[4pt]
How to locate the errors:\\
Read the numbered segments in order. For each one ask: given everything before it, is this step correct? List every segment that contains a genuine error, in the order in which they appear.\\[4pt]
Do NOT flag a segment merely because it is: an exploratory attempt the student later abandons or corrects; a longer route than necessary; or sloppy notation that does not change the mathematics. Flag only a step that is actually wrong.\\[4pt]
Rules you must follow:\\
1. List the erroneous segments in order of appearance.\\
2. For each one, quote verbatim the single sentence that carries the error.\\
3. For each one, say what is wrong and what to do instead. The student will resume writing after the last listed error, so phrase it as guidance they can act on. Two or three sentences per error.\\
4. NEVER reveal the final answer. Do not state it, do not write an expression that evaluates to it, do not say what it should be. Describe the method, not the result.\\
5. If every step looks correct to you and you cannot identify a genuine error, reply with SEGMENT: NONE. Do not guess {-}{-} a wrong location is worse than none.\\[4pt]
Reply in exactly this format, one block per error:\\
ERROR 1\\
SEGMENT: <index>\\
QUOTE: <the exact wrong sentence, copied verbatim>\\
FEEDBACK: <what is wrong and what to do instead>\\
ERROR 2\\
SEGMENT: <index>\\
QUOTE: <...>\\
FEEDBACK: <...>
\end{prompt}

\begin{prompt}[title={$\mathcal{P}_{\text{critic}}^{\text{len}}$ --- first paragraph of $\mathcal{P}_{\text{critic}}$ for a rollout that reaches the length limit}]
You are a meticulous math teacher. A student's solution ran on without ever reaching an answer. You are given the correct answer. Use it as a reference to trace the student's work forward and find the steps that sent the reasoning off track. If the work is still on a valid path and was merely slow, reply SEGMENT: NONE.
\end{prompt}

\begin{prompt}[title={$\mathcal{P}_{\text{critic-user}}$ --- user message of the critic}]
Problem:\\
\texttt{\{student prompt\}}\\[4pt]
Correct answer (reference only {-}{-} never reveal it): \texttt{\{reference answer\}}\\[4pt]
Student's solution, split into numbered segments:\\[4pt]
{[}0{]} \texttt{\{segment 0\}}\\[4pt]
{[}1{]} \texttt{\{segment 1\}}\\[4pt]
\dots
\end{prompt}

\begin{prompt}[title={$\mathcal{P}_{\text{fb}}$ --- text appended to the kept prefix of the rollout}]
\texttt{\{kept prefix\}}\\[4pt]
{[}Teacher feedback{]} \texttt{\{feedback\}}\\[4pt]
Using this feedback, continue the solution from here. Put your final answer in \verb|\boxed{}|.
\end{prompt}

\begin{prompt}[title={\texttt{\{feedback\}} in $\mathcal{P}_{\text{fb}}$ when the critic lists $k > 1$ errors; for $k = 1$, the lines before the last line are only \texttt{\{feedback 1\}}}]
Your solution contains \texttt{\{k\}} errors.\\
(1) In ``\texttt{\{quote 1\}}'': \texttt{\{feedback 1\}}\\
(2) In ``\texttt{\{quote 2\}}'': \texttt{\{feedback 2\}}\\
\dots\\
The correct final answer is \texttt{\{reference answer\}}.
\end{prompt}

Each quote is cut to $200$ characters.

\subsection{Prompts of the Ablations}
\label{app:prompts-ablation}

All variants in Table~\ref{tab:ablation} use the critic of \ours{}.
The variant without the answer omits the last line of the feedback, and its critic removes every listed error whose feedback contains the reference answer while the solution of the student does not.
The variants in (b) and (c) of Table~\ref{tab:ablation} change this variant, and the runs in Appendices~\ref{app:hyper} and~\ref{app:olmo} use the same feedback.
The variant without the critique keeps only the last line of the feedback, which states the reference answer, and the variant with generic feedback replaces the feedback by $\mathcal{P}_{\text{generic}}$.

\begin{prompt}[title={$\mathcal{P}_{\text{generic}}$ --- generic feedback}]
Your solution contains errors. Check the previous reasoning and redo it.
\end{prompt}

\subsection{Prompts of the Diagnosis}
\label{app:prompts-diag}

The diagnosis in Section~\ref{sec:critique-ability} uses the critic of \ours{} without the reference answer.
The first paragraph of $\mathcal{P}_{\text{critic}}$ is replaced by $\mathcal{P}_{\text{critic}}^{\text{diag}}$ for a complete solution and by $\mathcal{P}_{\text{critic}}^{\text{part}}$ for a truncated solution, and the line with the reference answer is removed from $\mathcal{P}_{\text{critic-user}}$.
The judge receives the system prompt $\mathcal{P}_{\text{judge}}$ and the user message $\mathcal{P}_{\text{judge-user}}$, which lists the errors that the critic claims.
The generic hint in Appendix~\ref{app:rewrite-analysis} replaces the feedback in $\mathcal{P}_{\text{fb}}$ by $\mathcal{P}_{\text{hint}}$.

\begin{prompt}[title={$\mathcal{P}_{\text{critic}}^{\text{diag}}$ --- first paragraph of $\mathcal{P}_{\text{critic}}$ in Section~\ref{sec:critique-ability}, complete solution}]
You are a meticulous math teacher. A student has submitted a complete solution whose FINAL ANSWER IS WRONG. Trace the student's work forward and find the steps that contain genuine errors.
\end{prompt}

\begin{prompt}[title={$\mathcal{P}_{\text{critic}}^{\text{part}}$ --- first paragraph of $\mathcal{P}_{\text{critic}}$ in Section~\ref{sec:critique-ability}, truncated solution}]
You are a meticulous math teacher. A student has written part of a solution, and the solution is not finished yet. Trace the student's work forward and find the steps that contain genuine errors. If every step so far is correct, reply SEGMENT: NONE.
\end{prompt}

\begin{prompt}[title={$\mathcal{P}_{\text{judge}}$ --- system prompt of the judge (Qwen3-8B)}]
You are an expert mathematics grader. You are given a math problem, its correct answer, a student's solution split into numbered segments, and a list of errors that a teacher claims to have found in the solution, in order of appearance. For every listed error, decide whether the claim is right.\\[4pt]
Labels:\\
NEW\_ERROR - the flagged step contains a genuine mistake of its own (a wrong formula, wrong logic, an arithmetic slip, a misread condition, an unjustified claim) that would still be wrong even if every earlier step were correct.\\
PROPAGATED - the reasoning in the flagged step is valid in itself; it is wrong only because it uses a value, expression or conclusion produced by an earlier mistake.\\
RESTATEMENT - the flagged step only restates or reports a wrong result that was already reached earlier (for example the final answer line or a summary), with no new reasoning.\\
NOT\_AN\_ERROR - the flagged step is actually correct.\\[4pt]
Reply with exactly one line per listed error and nothing else, in this format:\\
ERROR 1: \textless{}LABEL\textgreater{}\\
ERROR 2: \textless{}LABEL\textgreater{}
\end{prompt}

\begin{prompt}[title={$\mathcal{P}_{\text{judge-user}}$ --- user message of the judge; for a truncated solution, the third line reads ``Student's partial solution (not finished yet), split into numbered segments:''}]
Problem:\\
\texttt{\{student prompt\}}\\[4pt]
Correct answer: \texttt{\{reference answer\}}\\[4pt]
Student's solution, split into numbered segments:\\[4pt]
{[}0{]} \texttt{\{segment 0\}}\\[4pt]
{[}1{]} \texttt{\{segment 1\}}\\[4pt]
\dots\\[4pt]
Errors claimed by the teacher:\\[4pt]
ERROR 1\\
SEGMENT: \texttt{\{index\}}\\
QUOTE: \texttt{\{quote\}}\\
TEACHER'S NOTE: \texttt{\{feedback\}}\\[4pt]
\dots
\end{prompt}

\begin{prompt}[title={$\mathcal{P}_{\text{hint}}$ --- generic hint in Appendix~\ref{app:rewrite-analysis}}]
Your solution contains an error. Check it and continue.
\end{prompt}

\subsection{Prompts of the Baselines}
\label{app:prompts-baselines}

SFT, KD, GRPO, OPD, FastOPD, SKD, and RelayOPD use $\mathcal{P}_{\text{train}}$.
TRD lets the teacher rewrite one solution of the student per prompt with the system prompt $\mathcal{P}_{\text{TRD}}^{\text{sys}}$ and the user message $\mathcal{P}_{\text{TRD}}$.
OPSD-T gives the student $\mathcal{P}_{\text{train}}$ and scores its rollout with a teacher that receives $\mathcal{P}_{\text{OPSD}}$.
Critique-GRPO critiques an incorrect rollout as \ours{} does, including the reference answer at the end of the feedback, and the student then answers again in a new conversation with the system prompt $\mathcal{P}_{\text{CGRPO}}^{\text{sys}}$ and the user message $\mathcal{P}_{\text{CGRPO}}$.

\begin{prompt}[title={$\mathcal{P}_{\text{TRD}}^{\text{sys}}$ --- system prompt of the TRD rewrite}]
Please reason step by step, and put your final answer within \textbackslash{}boxed\{\}.
\end{prompt}

\begin{prompt}[title={$\mathcal{P}_{\text{TRD}}$ --- user message of the TRD rewrite}]
Your task is to rewrite your mathematical solution.\\[4pt]
**Problem:**\\
\texttt{\{student prompt\}}\\[4pt]
**Your Initial Solution:**\\
\texttt{\{initial solution\}}\\[4pt]
**Instructions:**\\
1. Preserve the overall structure and reasoning path of your original solution\\
2. Identify and fix errors in computation or logic\\
3. Keep correct intermediate steps and meaningful work\\
4. Output ONLY the rewritten solution
\end{prompt}

\begin{prompt}[title={$\mathcal{P}_{\text{OPSD}}$ --- user message of the teacher in OPSD-T}]
\texttt{\{student prompt\}}\\[4pt]
{[}Reference answer: \texttt{\{reference answer\}}{]}
\end{prompt}

\begin{prompt}[title={$\mathcal{P}_{\text{CGRPO}}^{\text{sys}}$ --- system prompt of the refinement in Critique-GRPO}]
You are a helpful assistant.
\end{prompt}

\begin{prompt}[title={$\mathcal{P}_{\text{CGRPO}}$ --- user message of the refinement in Critique-GRPO}]
Given the following inputs:\\
**Question**: \texttt{\{student prompt\}}\\
**Previous Solution**: \texttt{\{initial solution\}}\\
**Critique**: \texttt{\{feedback\}}\\[4pt]
Please re-answer by:\\
- Correcting potential errors identified in the critique, if they exist.\\
- Providing clear, step-by-step reasoning.\\
- Formatting the final answer as \verb|`\boxed{answer}`|.\\[4pt]
Ensure the revised solution addresses all issues raised in the critique.
\end{prompt}

\section{Details of the Diagnosis}
\label{app:diagnosis}

\subsection{Rollouts and Cut Points}

The diagnosis uses the Qwen3-1.7B student after $80$ steps of OPD with seed $42$
and the teacher of the main experiments.
The student generates one rollout for each of $900$ problems from
DAPO-Math-17k with temperature $1.0$, top-$p$ $1.0$, and at most $16{,}384$
tokens, and we grade each rollout with the reward function of training.
We randomly select $150$ correct and $150$ incorrect rollouts.
An incorrect rollout must end with a final answer, so we exclude rollouts that
reach the length limit.
We split each response into paragraphs at blank lines, as \ours{} does.
The cut points are the empty prefix and the $7$ paragraph boundaries that lie
closest to the positions $1/8, \dots, 7/8$ of the response.

\subsection{Continuations and Success Rates}

From each cut point, the teacher and the student each generate $16$
continuations of the prefix of the student within the same assistant turn.
Each model uses its own chat template, temperature $1.0$, top-$p$ $1.0$, and a
budget of $16{,}384$ tokens minus the length of the prefix.
$\Lambda^T(h)$ and $\Lambda^S(h)$ are the fractions of correct continuations of
the teacher and the student from the prefix $h$.
A problem is solvable by the teacher if the teacher succeeds in at least $8$ of
the $16$ continuations from the empty prefix.
All bands are $95\%$ bootstrap confidence intervals with $2{,}000$ resamples of
rollouts.

\subsection{Critique and Judgment}

For Figure~\ref{fig:critique}(b), the teacher critiques the truncated solution at
every cut point with $\mathcal{P}_{\text{critic}}^{\text{part}}$
(Appendix~\ref{app:prompts-diag}).
It decodes greedily for at most $2{,}048$ tokens, and we apply the filters of
training.
Qwen3-8B in thinking mode judges every listed error with the reference answer in
its context.
It labels an error as a new mistake, a mistake that follows from an earlier one,
a restatement of an earlier wrong result, or not an error, and we count every
label other than the last as real.
The judge samples with temperature $0.6$, top-$p$ $0.95$, and top-$k$ $20$ and
generates at most $16{,}384$ tokens.

\subsection{Rewrite Analysis}
\label{app:rewrite-analysis}

Figure~\ref{fig:rewrite-effect} uses the incorrect rollouts in which the success
rate of the teacher, estimated from its first $8$ continuations, is at least
$0.5$ at the empty prefix and falls below $0.5$ at a later cut point.
To enlarge this set, we add the $204$ remaining incorrect rollouts from the same
$900$ problems, which gives $163$ such rollouts.
The teacher critiques each complete solution with the prompts of \ours{}, and
$161$ critiques keep at least one error after filtering.
We cut each rollout as \ours{} does and draw $8$ continuations per condition.
At the cut point, the teacher continues with and without the critique, and the
student continues without the critique, with the critique, and with the generic
hint $\mathcal{P}_{\text{hint}}$.
At $25\%$, $50\%$, and $75\%$ of a rewrite that the student writes with the
critique, the teacher continues with the critique and the student continues
without it.
At the same positions of the discarded part of the original rollout, both
continue without the critique.

\subsection{Diagnosis with the Qwen3-0.6B Student}
\label{app:diag-06b}

\begin{figure}[t]
\centering
\includegraphics[width=\columnwidth]{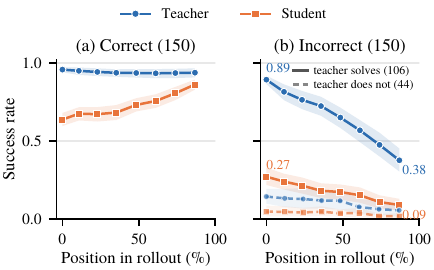}
\caption{Expected accuracy of the teacher and the Qwen3-0.6B student from prefixes of the student on (a) correct and (b) incorrect rollouts.}
\label{fig:failure-region-06b}
\end{figure}

\begin{figure}[t]
\centering
\includegraphics[width=\columnwidth]{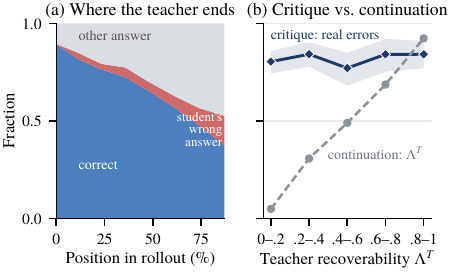}
\caption{(a) Outcomes of the continuations of the teacher and (b) fraction of the listed errors that are real, for the Qwen3-0.6B student.}
\label{fig:critique-06b}
\end{figure}

\begin{figure}[t]
\centering
\includegraphics[width=0.58\columnwidth]{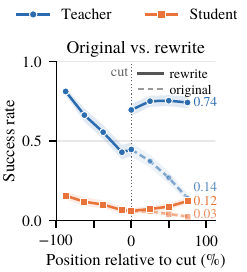}
\caption{Success rates of the teacher and the Qwen3-0.6B student around the cut point.}
\label{fig:rewrite-effect-06b}
\end{figure}

We repeat the diagnosis with the Qwen3-0.6B student after $40$ steps of OPD
with seed $42$, which is the checkpoint in Table~\ref{tab:main}, and keep the
teacher, the problems, and all other settings.
Of the $900$ rollouts, $270$ are correct, $593$ are incorrect, and $37$ reach the
length limit.
Figures~\ref{fig:failure-region-06b}, \ref{fig:critique-06b}, and
\ref{fig:rewrite-effect-06b} correspond to Figures~\ref{fig:failure-region},
\ref{fig:critique}, and \ref{fig:rewrite-effect}, and they show the same
pattern as the Qwen3-1.7B student.
On the $106$ incorrect rollouts from problems that the teacher solves,
$\Lambda^T$ falls from $0.89$ at the empty prefix to $0.38$ at the last cut
point, and the fraction of teacher continuations that end with the incorrect
answer of the student rises from $0.00$ to $0.15$.
Across the groups of $\Lambda^T$, whose means range from $0.05$ to $0.92$,
between $77\%$ and $84\%$ of the errors that the teacher lists are real.
For Figure~\ref{fig:rewrite-effect-06b}, we add the $443$ remaining incorrect
rollouts as in Appendix~\ref{app:diagnosis}, which gives $304$ rollouts in which
the success rate of the teacher falls below $0.5$, and $295$ critiques keep at
least one error.
After the cut point, the success rate of the teacher falls from $0.45$ to $0.14$
on the original rollout and stays between $0.70$ and $0.75$ on the rewrite, and
the success rate of the student rises from $0.06$ to $0.12$ on the rewrite.

\subsection{Occupancy Mismatch}
\label{app:occupancy}

\begin{figure*}[t]
\centering
\includegraphics[width=\textwidth]{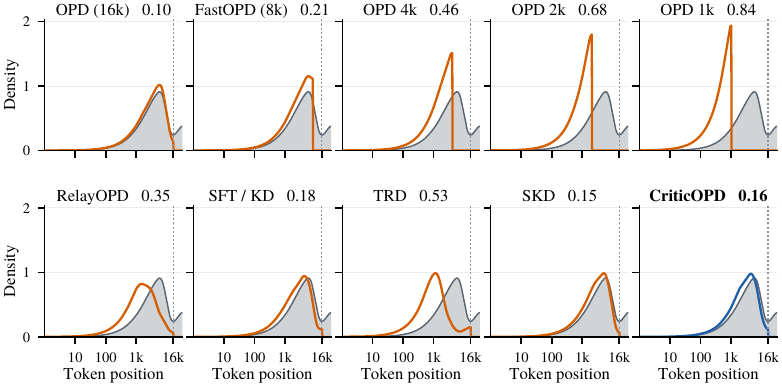}
\caption{Token positions of the prefixes that each method supervises (colored) and of the inference prefixes (gray), with the total variation distance in each title. The dotted line marks the training limit of $16{,}384$ tokens.}
\label{fig:mismatch-positions}
\end{figure*}

\begin{figure*}[t]
\centering
\includegraphics[width=\textwidth]{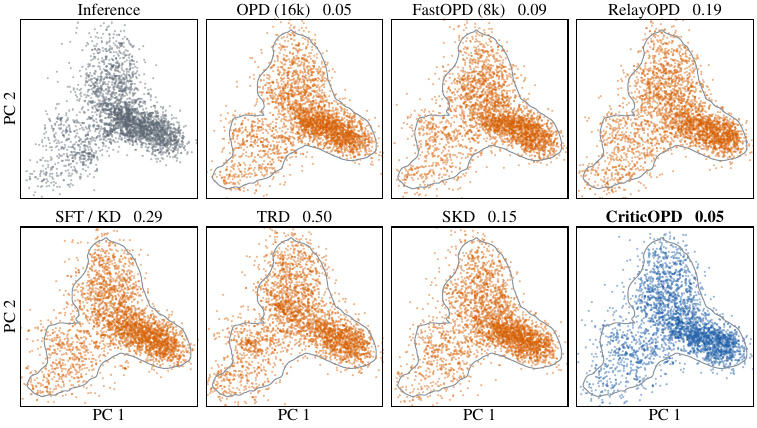}
\caption{Hidden states of the student on the prefixes that each method supervises, projected onto the first two principal components of the inference states. The contour encloses $95\%$ of the inference states, and each title gives the classifier bound on the total variation distance.}
\label{fig:mismatch-hidden}
\end{figure*}

The measurement in Section~\ref{sec:experiments-why} uses the first $256$ problems in the training order of DAPO-Math-17k with seed $42$ and the student of Section~\ref{sec:diagnosis}.
For $d_\theta$, the student generates four rollouts per problem with temperature $1.0$, top-$p$ $1.0$, and at most $32{,}768$ tokens, and every generated token contributes one position.
For $d_m$, CriticOPD, RelayOPD, and SKD each run one training step on these problems from the same student, and every token that receives a loss contributes one position.
For OPD, we take all positions of the original rollouts in the step of CriticOPD, which the student generates with the training limit of $16{,}384$ tokens.
FastOPD differs from OPD only in the length limit, so we take the first $8{,}192$ positions of these rollouts, and the first $1{,}024$, $2{,}048$, and $4{,}096$ positions for the shorter limits.
SFT and KD train on the same teacher solutions of these problems, and TRD trains on the solutions that the teacher rewrites from rollouts of the student, so every response token of these data contributes one position.
We split the positions from $1$ to $32{,}768$ into $48$ intervals of equal width on a logarithmic scale and merge the intervals whose rounded boundaries coincide, which leaves $45$ intervals.
We compute the total variation distance between the two position distributions over these intervals.
The curves in Figure~\ref{fig:mismatch-positions} are kernel density estimates over the logarithm of the position and serve only for display.

Figure~\ref{fig:mismatch-positions} shows the position distributions of all methods.
A shorter length limit moves the supervision to earlier positions, and the distance grows from $0.10$ for OPD to $0.21$, $0.46$, $0.68$, and $0.84$ for limits of $8{,}192$, $4{,}096$, $2{,}048$, and $1{,}024$ tokens.
RelayOPD and TRD supervise shorter sequences than the student generates, with distances of $0.35$ and $0.53$.
SFT, KD, and SKD stay close to inference in position, with distances of $0.18$ and $0.15$.
At inference, $9.3\%$ of the tokens lie after position $16{,}384$, which no method with a training limit of $16{,}384$ tokens supervises.

The position does not reflect the content of a prefix, so we also compare the hidden states of the student, which are the inputs to its output layer at the prefixes.
We sample $20{,}000$ states per method and $30{,}000$ states for inference uniformly over tokens, subtract the mean inference state of each problem, and divide each dimension by its standard deviation over the inference states.
A logistic regression separates the states of a method from an equal number of inference states, with five folds that split the problems.
If its held-out accuracy is $a$, then $2a-1$ is a lower bound on the total variation distance between the two state distributions.
Two disjoint halves of the inference rollouts give $-0.02$, which is the noise level.
Figure~\ref{fig:mismatch-hidden} projects the states onto the first two principal components of the inference states.
CriticOPD has a distance of $0.05$, the same as OPD and lower than FastOPD ($0.09$), SKD ($0.15$), and RelayOPD ($0.19$).
SFT and KD ($0.29$) and TRD ($0.50$) train on text that the teacher writes or rewrites and have the largest distances in hidden states, although SFT and KD stay close to inference in position.

\section{Experimental Details}
\label{app:setup}

\subsection{Training}

\begin{table}[h]
\centering
\caption{\textbf{Hyperparameters} of all methods.}
\label{tab:hyper}
\scriptsize
\setlength{\tabcolsep}{4pt}
\renewcommand{\arraystretch}{1.08}
\begin{tabular}{ll}
\toprule
\textbf{Setting} & \textbf{Value} \\
\midrule
\multicolumn{2}{l}{\textit{Training}} \\
Prompts per step & $128$ \\
Rollouts per prompt & $1$ (GRPO: $8$) \\
Optimizer updates per step & $1$ \\
Learning rate & $10^{-6}$, constant, no warmup \\
Maximum prompt length & $2{,}048$ \\
Maximum response length & $16{,}384$ (FastOPD: $8{,}192$) \\
Rollout temperature, top-$p$ & $1.0$, $1.0$ \\
Checkpoint interval & $20$ steps \\
GPUs & $4$ ($2$ student, $2$ teacher) \\
\midrule
\multicolumn{2}{l}{\textit{Critic of \ours{}}} \\
Decoding & greedy \\
Maximum critique length & $2{,}048$ \\
Number of listed errors & all \\
\midrule
\multicolumn{2}{l}{\textit{Evaluation}} \\
Temperature, top-$p$ & $1.0$, $1.0$ \\
Maximum new tokens & $32{,}768$ \\
Samples per problem & $32$ (AIME, AMC), $8$ (others) \\
\bottomrule
\end{tabular}
\end{table}

Table~\ref{tab:hyper} lists the shared hyperparameters.
OPD optimizes the sampled reverse-KL objective of Equation~\eqref{eq:adv} with
the policy gradient and applies no KL penalty.
One epoch over DAPO-Math-17k takes $139$ steps.

\subsection{Software and Hardware}
\label{app:software}

We implement all methods on the code of RelayOPD \citep{xu2026relay}, which
builds on \texttt{verl} 0.9.0.dev \citep{sheng2025hybridflow}, and the released
code fixes the commit of this code base.
The software consists of Python 3.12, PyTorch 2.11 with CUDA 13.0,
Transformers 5.15, and Ray 2.57.
vLLM 0.21 \citep{kwon2023vllm} generates the rollouts and the critiques, and a
separate vLLM server computes the log-probabilities of the teacher on the
training sequences.
We grade mathematical answers with Math-Verify 0.9.0 and code with EvalPlus
0.3.1.
Each training run uses one node with four NVIDIA A100-SXM4-80GB GPUs, two for
the student and two for the teacher.
The seed-$42$ runs of OPD, FastOPD, and RelayOPD on Qwen3-1.7B use four GPUs
for each model with the same batch size, and GRPO and SFT, which use no
teacher during training, use all four GPUs for the student.
Runs on Qwen3-0.6B use A100, H100, H200, or B200 GPUs, and each evaluation job
uses one GPU.

\subsection{Configuration of \ours{}}
\label{app:config}

Algorithm~\ref{alg:criticopd} summarizes one training step of \ours{}.
The critic is the teacher.
If a critique reaches the length limit, we discard its last item, which is
incomplete.
The feedback lists the remaining items and ends with the reference answer
(Appendix~\ref{app:prompts-method}).
Rollouts that reach the response limit are also sent to the critic, with a
prompt that tells the critic the solution is unfinished
(Appendix~\ref{app:prompts-method}).
The rewrite uses the response budget that remains after the kept prefix.
We train \ours{} for $40$ steps on Qwen3-1.7B and for $80$ steps on Qwen3-0.6B,
and select the step as for all methods (Appendix~\ref{app:selection}).

\begin{algorithm}[t]
\caption{CriticOPD Training}
\label{alg:criticopd}
\begin{algorithmic}[1]
\Require Student $\pi_\theta$, teacher $\pi_T$,
batch $\{(x,a)\}$
\State $\mathcal{L}_{\mathrm{ours}} \gets 0$
\For{each $(x,a)$ in the batch}
    \State Sample $y \sim \pi_\theta(\cdot\mid x)$
    \If{$y$ is incorrect or truncated}
        \State Generate critique $c$ using $\pi_T$ given $(x,y,a)$
        \State Discard incomplete error items from $c$
    \EndIf
    \If{a valid critique is available}
        \State $\tau \gets$ first position after the last critiqued paragraph
        \State $\tilde{y}_{<\tau} \gets y_{<\tau}$
        \State Sample $\tilde{y}_{\geq\tau}
        \sim \pi_\theta(\cdot\mid h_\tau,c)$
        \State $\ell \gets \mathcal{L}_{\mathrm{fb}}(\tilde{y};c)$
    \Else
        \State $\ell \gets \mathcal{L}_{\mathrm{OPD}}(y)$
    \EndIf
    \State $\mathcal{L}_{\mathrm{ours}}
    \gets \mathcal{L}_{\mathrm{ours}}+\ell$
\EndFor
\State Update $\theta$ using $\mathcal{L}_{\mathrm{ours}}$
\end{algorithmic}
\end{algorithm}

\begin{figure*}[t]
\centering
\includegraphics[width=\textwidth]{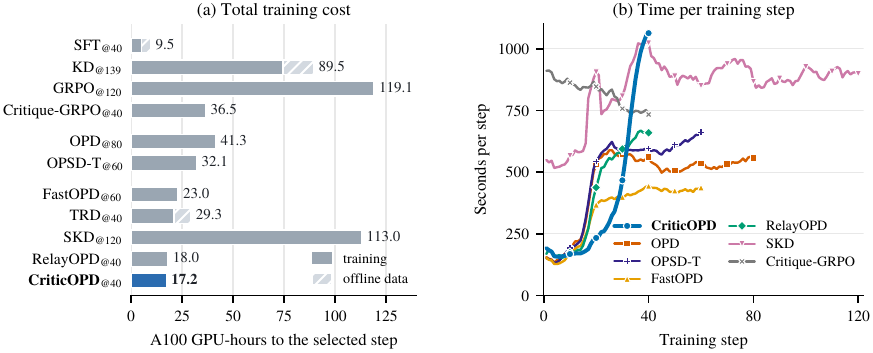}
\caption{Training cost of the Qwen3-1.7B students. (a) A100 GPU-hours until the
selected step, including the generation of offline data. (b) Seconds per
training step, averaged over five steps.}
\label{fig:cost}
\end{figure*}

\subsection{Baselines}

SFT trains the student with cross-entropy on one solution of the teacher per
prompt, which the teacher samples with temperature $1.0$ and which we do not
filter by correctness.
KD trains on the same solutions with the forward KL divergence on the $128$
most probable tokens of the teacher.
GRPO uses a binary correctness reward and no teacher.
FastOPD is OPD with a maximum response length of $8{,}192$ tokens during
training.
TRD lets the teacher rewrite one solution of the student per prompt and fix its
errors, and it trains the student on the rewrites with the objective of KD.
SKD generates rollouts by speculative decoding: the teacher accepts a draft
token of the student if the token is among its $5$ most probable tokens and
otherwise samples a replacement, and the student is trained with the objective
of KD.
RelayOPD lets the teacher take over for at most $3$ paragraphs or $256$ tokens
when the teacher predicts a reflection token that is outside the $5$ most
probable tokens of the student.
It allows at most two takeovers, ends the rollout after the second one, and
applies the OPD objective to all tokens.
OPSD-T trains as OPD, except that the teacher scores each rollout with the
reference answer after the problem ($\mathcal{P}_{\text{OPSD}}$ in
Appendix~\ref{app:prompts-baselines}).
The student generates and is trained without the reference answer, and no
rollout is critiqued or rewritten.
Critique-GRPO follows its official implementation and uses only the outcome
reward.
The student samples $7$ rollouts per prompt.
If all of them are incorrect, the critic of \ours{} critiques each of them with
the feedback ending with the reference answer, and the student answers again in
a new conversation with $\mathcal{P}_{\text{CGRPO}}$.
One of these refinements, a correct one if any, joins the group as an eighth
response.
The advantage is the reward minus the mean reward of the group, without the
normalization by the standard deviation that GRPO uses.
Tokens of the initial rollouts use the importance ratio without clipping, the
tokens of the refinement are weighted by $p/(p+0.1)$, where $p$ is their
probability under the student, and the entropy coefficient is $0.001$.

\subsection{Evaluation}

The benchmarks contain $30$ (AIME 2024), $30$ (AIME 2025), $83$ (AMC 2023),
$500$ (MATH500), $675$ (OlympiadBench), and $272$ (Minerva Math) problems.
The students generate in non-thinking mode, and Avg is the macro average over
the six benchmarks.
For each method, we select one step with seed $42$
(Appendix~\ref{app:selection}), train seeds $43$ and $44$ to the same step, and
report the mean and the sample standard deviation over the three seeds.
The teacher and the initial students are evaluated once.

\subsection{Checkpoint Selection}
\label{app:selection}

We select checkpoints on a validation set that does not overlap with the
training data or the six test benchmarks.
We call a problem contaminated if at least half of its $13$-grams appear in
DAPO-Math-17k or in a test benchmark, or if its normalized text equals that of
such a problem.
A single shared $13$-gram is not enough, because common phrases such as ``where
$m$ and $n$ are relatively prime positive integers. Find $m+n$'' already contain
more than $13$ words.
The validation set has two parts.
MATH-val starts from the $4{,}500$ test problems of MATH outside MATH500, of
which $1{,}304$ are contaminated, $1{,}284$ of them by DAPO-Math-17k, and we
sample $500$ of the remaining $3{,}196$ problems uniformly with seed $0$.
Comp-val contains all $153$ problems of four competitions from 2025 in MathArena
\citep{balunovic2025matharena}, namely BRUMO, CMIMC, HMMT February, and SMT, and
none of them is contaminated.
We do not use AIME 2022 and 2023, because $54$ of their $60$ problems appear in
DAPO-Math-17k.
The validation set uses the prompt $\mathcal{P}_{\text{eval}}$, the chat
template, the stop tokens, the grader, and the sampling parameters of the test
benchmarks, with up to $32{,}768$ new tokens, and it draws $4$ samples per
problem on MATH-val and $8$ on Comp-val.
The accuracy of each part averages the fraction of correct samples over its
problems, and the validation score is the mean of the accuracies of the two
parts.

The candidates of each method are the checkpoints at every multiple of $20$
steps and at the last step of training if it is not a multiple of $20$.
We select the candidate with the highest validation score, and the earlier step
if two candidates tie.
For Qwen3-1.7B and Qwen3-0.6B, we select once for each student and method with
the checkpoints of seed $42$ and use the selected step for all seeds.
For the OLMo-3 student, we make no selection and reuse the steps selected for
Qwen3-1.7B, with the exceptions that Appendix~\ref{app:olmo} describes.
The test results do not enter the selection.

\begin{table*}[t]
\centering
\caption{
\textbf{Results with OLMo-3 models} on six mathematical reasoning benchmarks with seed $42$.
The subscript after each method gives the step, and bold marks the best student.
}
\label{tab:olmo}

\scriptsize
\setlength{\tabcolsep}{3pt}
\renewcommand{\arraystretch}{1.05}

\begin{tabularx}{\textwidth}{l*{7}{>{\centering\arraybackslash}X}}
\toprule
\textbf{Method}
& \textbf{AIME24}
& \textbf{AIME25}
& \textbf{AMC23}
& \textbf{MATH500}
& \textbf{OlympiadBench}
& \textbf{Minerva}
& \textbf{Avg}
\\
\midrule

Teacher
& 73.54 & 60.62 & 92.24 & 96.22 & 77.35 & 47.43
& 74.57 \\

\midrule
\multicolumn{8}{c}{\textit{Student: Olmo-3-7B-Instruct-SFT, teacher: Olmo-3.1-32B-Instruct}} \\
\midrule

Student
& 6.15 & 4.17 & 30.50 & 64.80 & 30.09 & 29.09
& 27.47 \\

\arrayrulecolor{black!35}\specialrule{0.35pt}{1.2pt}{1.2pt}\arrayrulecolor{black}

SFT$_{@40}$
& 31.56 & 26.15 & 59.45 & 79.50 & 51.91 & 31.66
& 46.70 \\

KD$_{@80}$
& 35.42 & 30.83 & 72.33 & 86.83 & 59.48 & 33.27
& 53.03 \\

GRPO$_{@120}$
& 10.00 & 7.60 & 37.46 & 69.85 & 36.28 & 30.70
& 31.98 \\

Critique-GRPO$_{@40}$
& 6.56 & 4.90 & 31.40 & 65.67 & 31.15 & 29.32
& 28.17 \\

\arrayrulecolor{black!35}\specialrule{0.35pt}{1.2pt}{1.2pt}\arrayrulecolor{black}

OPD$_{@60}$
& 41.04 & 31.98 & 71.54 & 87.58 & 60.61 & 35.39
& 54.69 \\

OPSD-T$_{@60}$
& 41.46 & 34.90 & 72.29 & 88.40 & 61.80 & 34.42
& 55.54 \\

\arrayrulecolor{black!35}\specialrule{0.35pt}{1.2pt}{1.2pt}\arrayrulecolor{black}

FastOPD$_{@60}$
& 38.23 & 33.33 & 71.05 & 88.05 & 60.65 & 35.52
& 54.47 \\

TRD$_{@40}$
& 27.50 & 24.48 & 56.25 & 79.55 & 50.70 & 31.66
& 45.02 \\

SKD$_{@60}$
& 38.82 & 29.33 & 70.82 & 86.61 & 59.30 & 33.78
& 53.11 \\

RelayOPD$_{@40}$
& 32.60 & 27.08 & 67.47 & 84.67 & 57.22 & 33.55
& 50.43 \\

\rowcolor{blue!8}
\textbf{\ours{}}$_{@60}$
& \textbf{43.44} & \textbf{37.81} & \textbf{74.70}
& \textbf{89.08} & \textbf{62.33} & \textbf{36.02}
& \textbf{57.23} \\

\rowcolor{blue!8}
\textcolor{green!50!black}{$\Delta$ vs OPD}
& \textcolor{green!50!black}{+2.40}
& \textcolor{green!50!black}{+5.83}
& \textcolor{green!50!black}{+3.16}
& \textcolor{green!50!black}{+1.50}
& \textcolor{green!50!black}{+1.72}
& \textcolor{green!50!black}{+0.63}
& \textcolor{green!50!black}{+2.54} \\

\bottomrule
\end{tabularx}
\end{table*}

\section{Computational Cost}
\label{app:cost}

Figure~\ref{fig:cost} compares the training cost of all methods on the
Qwen3-1.7B student.
We measure the time of every training step under seeds $43$ and $44$, in which
all methods run on four A100 GPUs, and average the two seeds.
For SFT, KD, and TRD, we add the time to generate the offline data on A100 GPUs.
The time per step of \ours{} grows from about $170$ seconds over the first $10$
steps to about $870$ seconds over steps $31$ to $40$, compared with $160$ and
$560$ seconds for OPD, because the responses of the student grow longer and
incorrect rollouts receive a critique and a rewrite.
Over the same $40$ steps, \ours{} uses $0.96$ times the GPU-hours of OPD.
\ours{} reaches its selected step after $40$ steps, so its total cost of
$17.2$ GPU-hours is $0.42$ times that of OPD at step $80$.
GRPO and Critique-GRPO, which sample $8$ rollouts per prompt, take about $890$
and $820$ seconds per step.

\begin{table*}[t]
\centering
\caption{
\textbf{Results outside mathematics} for Qwen3-1.7B. Code uses EvalPlus
pass@1 on the extended tests.
}
\label{tab:ood}

\scriptsize
\setlength{\tabcolsep}{3pt}
\renewcommand{\arraystretch}{1.05}

\begin{tabularx}{\textwidth}{l*{7}{>{\centering\arraybackslash}X}}
\toprule
\textbf{Method}
& \textbf{MMLU-Pro}
& \textbf{MMLU}
& \textbf{ARC-C}
& \textbf{OpenBookQA}
& \textbf{HumanEval+}
& \textbf{MBPP+}
& \textbf{Avg}
\\
\midrule

Teacher
& 69.75 & 80.09 & 94.94 & 93.95 & 82.00 & 74.70
& 82.57 \\

Student
& 38.10 & 55.24 & 77.69 & 74.10 & 62.50 & 56.50
& 60.69 \\

\midrule

SFT
& 50.80 & 63.29 & 84.41 & 79.70 & 63.10 & 55.40
& 66.12 \\
KD
& 50.85 & 63.78 & 84.28 & 79.30 & 64.00 & 57.10
& 66.55 \\
GRPO
& 51.41 & 61.49 & 81.68 & 78.75 & 63.70 & 58.50
& 65.92 \\
\arrayrulecolor{black!35}\specialrule{0.35pt}{1.2pt}{1.2pt}\arrayrulecolor{black}
OPD
& 56.61 & 67.14 & 87.07 & 83.80 & 66.30 & 61.90
& 70.47 \\
\arrayrulecolor{black!35}\specialrule{0.35pt}{1.2pt}{1.2pt}\arrayrulecolor{black}
FastOPD
& 56.14 & 66.27 & 85.52 & 83.40 & 66.50 & 61.20
& 69.84 \\
TRD
& 47.44 & 61.22 & 81.08 & 75.45 & 62.80 & 57.20
& 64.20 \\
SKD
& 53.84 & 65.87 & 85.90 & 81.65 & 63.70 & 59.90
& 68.48 \\
RelayOPD
& 56.04 & 66.40 & 85.90 & 83.20 & 66.90 & 60.40
& 69.81 \\
\rowcolor{blue!8}
\textbf{\ours{}}
& 57.31 & 67.88 & 87.73 & 84.90 & 65.90 & 60.40
& 70.69 \\

\bottomrule
\end{tabularx}
\end{table*}

\section{Additional Results}
\label{app:results}

\subsection{Results with OLMo-3 Models}
\label{app:olmo}

Table~\ref{tab:olmo} repeats the comparison with models from another family.
The student is Olmo-3-7B-Instruct-SFT, and the teacher is Olmo-3.1-32B-Instruct, which shares the tokenizer of the student.
All methods use the training recipe and the evaluation protocol of Table~\ref{tab:main} with seed $42$.
Each method is trained to the step selected for Qwen3-1.7B, except that OPD and \ours{} are both evaluated at step $60$, and that we stop KD and SKD early, at steps $80$ and $60$ instead of $139$ and $120$.
The critique of \ours{} here does not end with the reference answer.
\ours{} raises the average of OPD from $54.69$ to $57.23$.
As with the Qwen3 students, the gains are largest on the competition benchmarks, with $+5.83$ on AIME 2025 and $+3.16$ on AMC 2023.

\begin{table*}[t]
\centering
\caption{\textbf{Per-benchmark results of the ablations} on Qwen3-1.7B with
seed $42$. The rows follow Table~\ref{tab:ablation}.}
\label{tab:ablation-full}
\scriptsize
\setlength{\tabcolsep}{3pt}
\renewcommand{\arraystretch}{1.05}
\begin{tabularx}{\textwidth}{l *{7}{>{\centering\arraybackslash}X}}
\toprule
\textbf{Variant} & \textbf{AIME24} & \textbf{AIME25} & \textbf{AMC23} & \textbf{MATH500} & \textbf{Olympiad} & \textbf{Minerva} & \textbf{Avg} \\
\midrule
\rowcolor{blue!8}
\ours{} & 40.52 & 29.06 & 70.18 & 89.40 & 60.59 & 37.64 & 54.57 \\
\midrule
\multicolumn{8}{@{}l}{\textit{(a) Content of the feedback}} \\
\quad \textit{w/o} answer & 41.56 & 27.60 & 67.43 & 88.75 & 59.04 & 37.91 & 53.72 \\
\quad \textit{w/o} critique & 34.48 & 21.96 & 60.33 & 85.55 & 54.83 & 34.89 & 48.67 \\
\quad critique + answer $\rightarrow$ generic & 40.00 & 27.71 & 67.39 & 88.05 & 58.91 & 36.63 & 53.11 \\
\midrule
\multicolumn{8}{@{}l}{\textit{(b) Writer of the suffix}} \\
\quad student $\rightarrow$ teacher & 38.12 & 28.23 & 68.45 & 87.48 & 59.37 & 36.72 & 53.06 \\
\midrule
\multicolumn{8}{@{}l}{\textit{(c) Scoring of the suffix}} \\
\quad \textit{w/o} feedback & 38.85 & 27.19 & 67.09 & 87.72 & 58.41 & 35.94 & 52.53 \\
\midrule
OPD & 34.69 & 22.71 & 63.33 & 86.70 & 56.31 & 37.32 & 50.18 \\
\bottomrule
\end{tabularx}
\end{table*}

\subsection{Results Outside Mathematics}
\label{app:ood}

Table~\ref{tab:ood} reports the results of the Qwen3-1.7B students outside mathematics.
We evaluate on MMLU-Pro, MMLU, ARC-Challenge, OpenBookQA, HumanEval+, and MBPP+
with $4$ samples per problem and a budget of $32{,}768$ tokens.
For MMLU and MMLU-Pro, we use subsets of $2{,}000$ questions that keep the
proportion of each subject.
We grade multiple-choice answers by the option letter in the final box and
report EvalPlus pass@1 on the extended tests for code.

\subsection{Per-Benchmark Ablation Results}
\label{app:ablation}

Table~\ref{tab:ablation-full} reports the six benchmarks for every variant in
Table~\ref{tab:ablation}, all at step $40$ except OPD at step $80$.

\subsection{Sensitivity to Hyperparameters}
\label{app:hyper}

\begin{figure}[H]
\centering
\includegraphics[width=\columnwidth]{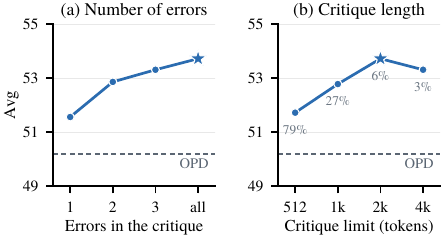}
\caption{Sensitivity of \ours{} on the Qwen3-1.7B student with seed $42$. The
star marks the default, and the numbers in (b) give the share of truncated
critiques.}
\label{fig:hyper}
\end{figure}

\ours{} adds two hyperparameters to OPD: the number $k$ of listed errors that
the feedback contains, and the length limit of the critique.
The rewrite starts after the $k$-th error.
Figure~\ref{fig:hyper} varies each hyperparameter and keeps the other at its
default, which uses all listed errors and a limit of $2{,}048$ tokens.
The average rises with $k$, from $51.56$ with one error to $53.31$ with three
and $53.72$ with all errors, so the feedback on later errors still helps.
The length limit matters until most critiques fit: limits of $512$ and
$1{,}024$ tokens cut off $79\%$ and $27\%$ of the critiques and give $51.72$
and $52.78$, whereas limits of $2{,}048$ and $4{,}096$ tokens cut off $6\%$ and
$3\%$ and give similar averages ($53.72$ and $53.31$).

\begin{figure*}[t]
\centering
\includegraphics[width=\textwidth]{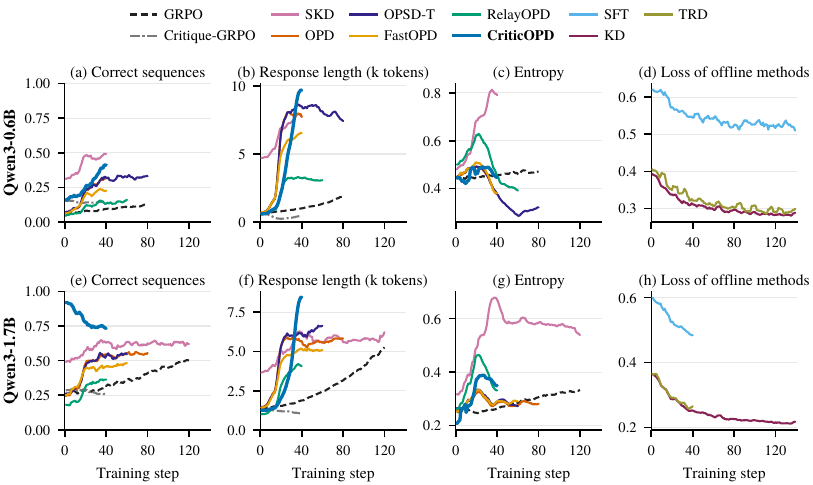}
\caption{Training dynamics, averaged over seeds and smoothed over five steps.}
\label{fig:dynamics}
\end{figure*}

\subsection{Training Dynamics}
\label{app:dynamics}

Figure~\ref{fig:dynamics} shows the training dynamics of all methods.
Panels (a) and (e) report the fraction of training sequences that reach the
reference answer.
These sequences contain the tokens that the teacher samples in SKD, the
takeovers of the teacher in RelayOPD, the refinements in Critique-GRPO, and the
rewrites after the critique in \ours{}, so this fraction is not the accuracy of
the student alone.
The entropy of Critique-GRPO is not shown, because its loss sums over the tokens
and divides by a constant length, so its logged entropy is not a mean over
tokens.
Panels (d) and (h) report the training loss of the methods that train on
offline data, which is the cross-entropy for SFT and the forward KL divergence
for KD and TRD.
The training logs of seed $42$ are missing for SFT, KD, and TRD on Qwen3-0.6B,
so these curves average seeds $43$ and $44$.
At both scales, the training sequences of \ours{} become longer than those of
OPD, which agrees with the longer responses at evaluation
(Section~\ref{sec:experiments-why}).

\subsection{Generation Length}
\label{app:budget}

\begin{figure}[!t]
\centering
\includegraphics[width=\columnwidth]{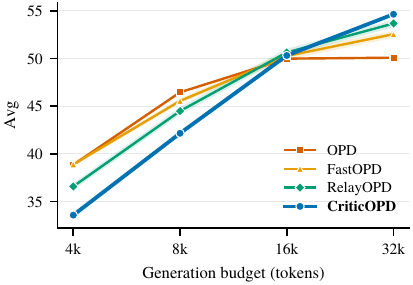}
\caption{Average over six benchmarks when responses longer than the budget
count as incorrect, over three seeds of Qwen3-1.7B.}
\label{fig:budget}
\end{figure}

The student of \ours{} writes longer responses than the student of OPD, with
$10.2$k tokens per response on average, compared with $5.3$k.
Figure~\ref{fig:budget} evaluates the Qwen3-1.7B students under generation
budgets from $4$k to $32$k tokens.
We reuse the responses with the budget of $32{,}768$ tokens and count a response
as incorrect if it is longer than the budget.
\ours{} falls below OPD at $4$k and $8$k tokens and exceeds it at $16$k and
$32$k.

\begin{table}[!t]
\centering
\caption{\textbf{Length-controlled comparison} on Qwen3-1.7B, averaged over
three seeds.}
\label{tab:length}
\footnotesize
\setlength{\tabcolsep}{4pt}
\begin{tabular}{@{}lccc@{}}
\toprule
\textbf{Method} & \textbf{AIME length} & \textbf{Avg@16k} & \textbf{Avg@32k} \\
\midrule
OPD & $8{,}487$ & 49.99 & 50.08 \\
FastOPD & $11{,}910$ & 50.27 & 52.55 \\
\midrule
\rowcolor{blue!8}
\ours{} & $12{,}457$ & 50.32 & 54.65 \\
\quad scoring \textit{w/o} feedback & $12{,}065$ & 50.62 & 52.26 \\
\quad critique $\rightarrow$ generic & $12{,}051$ & 50.86 & 52.87 \\
\bottomrule
\end{tabular}
\end{table}

Longer responses alone do not explain the gain.
Table~\ref{tab:length} compares variants that share the rewriting of \ours{} and
produce responses of about the same median length on AIME.
At the 32k generation budget, when the teacher scores the rewrite without the critique, the average is
$2.39$ points lower than that of \ours{} with the same feedback, and generic
feedback lowers it by $1.78$ points.
FastOPD also writes responses of a similar length and is $2.10$ point lower.
At the same response length, the critique in the context of the teacher
therefore still improves the student.

\end{document}